\documentclass{article}

\usepackage[preprint]{neurips_2022}

\usepackage{amsmath,amsfonts,bm}

\def\eqref#1{equation~\ref{#1}}
\def\1{\bm{1}}

\DeclareMathAlphabet{\mathsfit}{\encodingdefault}{\sfdefault}{m}{sl}
\SetMathAlphabet{\mathsfit}{bold}{\encodingdefault}{\sfdefault}{bx}{n}

\usepackage[utf8]{inputenc} % allow utf-8 input
\usepackage[T1]{fontenc}    % use 8-bit T1 fonts
\usepackage{hyperref}       % hyperlinks
\usepackage{url}            % simple URL typesetting
\usepackage{booktabs}       % professional-quality tables
\usepackage{amsfonts}       % blackboard math symbols
\usepackage{nicefrac}       % compact symbols for 1/2, etc.
\usepackage{microtype}      % microtypography
\usepackage{xcolor}         % colors

\usepackage[shortlabels]{enumitem}
\usepackage{bm}
\usepackage{mathtools}
\usepackage{subcaption}
\usepackage{soul}           % For highlighting text with \hl{...}
\usepackage{enumitem}       % enumerate by (i), (ii),...
\usepackage{bigints}        % for big integral symbols
\usepackage{hyperref}
\usepackage{amsmath,amsthm,amssymb}

\usepackage{placeins} %Float Barrier

\usepackage{algorithm}
\usepackage{algpseudocode}

\newcommand{\bs}[1]{\bm{#1}}

\newcommand{\txi}[1]{t^{(#1)}, \bs x^{(#1)}}

\newcommand{\molecDist}{\Psi}

\newcommand{\unibas}{%
University of Basel%\\
}

\title{Mesh-free Eulerian Physics-Informed Neural Networks}

\author{%
  Fabricio Arend Torres, Marcello M. Negri, Monika Nagy-Huber, Maxim Samarin \& Volker Roth\\
  \unibas \\
  \texttt{\{fabricio.arendtorres, marcellomassimo.negri}\\ 
   \texttt{ monika.nagy, maxim.samarin, volker.roth\}@unibas.ch} \\
  \\
}
\begin{document}

\maketitle

%%%%%%%%%%%%%%%%%%%%%%%%%%%%%%%%%%%%%%%%%%%%%%%%%%%%%%%%%%%%%%%%%%%%%%%%
%%%%%%%%%%%%%%%%%%%%%%%%%%%%%%%%%%%%%%%%%%%%%%%%%%%%%%%%%%%%%%%%%%%%%%%%
\begin{abstract}
Physics-informed Neural Networks (PINNs) have recently emerged as a principled way to include prior physical knowledge in form of partial differential equations (PDEs) into neural networks.
Although PINNs are generally viewed as mesh-free, current approaches still rely on collocation points 
within a bounded region, even in settings with spatially sparse signals.
Furthermore, if the boundaries are not known, the selection of such a region is difficult and often results in a large proportion of collocation points being selected in areas of low relevance. 
To resolve this severe drawback of current methods, we present a mesh-free and adaptive approach termed particle-density PINN (pdPINN), which is inspired by the microscopic viewpoint of fluid dynamics.
The method is based on the Eulerian formulation and, different from classical mesh-free method, does not require the introduction of Lagrangian updates.
We propose to sample directly from the distribution over the particle positions, eliminating the need to introduce boundaries while adaptively focusing on the most relevant regions.
This is achieved by interpreting a non-negative physical quantity (such as the density or temperature) 
as an unnormalized probability distribution from which we sample with dynamic Monte Carlo methods.
The proposed method leads to higher sample efficiency and improved performance of PINNs.
These advantages are demonstrated on various experiments based on the continuity equations, Fokker-Planck equations, and the heat equation.
\end{abstract}

\section{Introduction}
% just edit the file `subfiles/introduction.tex`
% \input{file} basically directly copies the text from the file into this main-file - there are no format changes.
% Splitting up the sections makes it easier to navigate.
\label{sec:introduction}
Many phenomena in physics are commonly described by partial differential equations (PDEs) which give rise to complex dynamical systems but often lack tractable analytical solutions. 
Important examples can be found for instance in fluid dynamics %and in modeling of compressible fluids 
with typical applications in the design of gas and steam turbines \citep{oosthuizen2013introduction},
as well as modeling the collective motion of self-driven particles \citep{marchetti2013hydrodynamics} such as flocks of birds or bacteria colonies \citep{szabo2006phase, nussbaumer2021quantifying}.
Despite the relevant progress in establishing numerical PDE solvers, such as finite element and finite volume methods, the seamless incorporation of data remains an open problem \citep{freitag2020numerical}. %, with
%Classical numerical data assimilation approaches can be divided into two types of methods \citep{xun2013parameter}, both of which are, however, very time consuming. 
To fill this gap, Physics-informed Neural Networks (PINNs) %, also referred to as Implicit Neural Representations \citep{sitzmann2020implicit}, 
have emerged as an attractive alternative to classical methods for data-based forward and inverse solving of PDEs.

The general idea of PINNs is to use the expressive power of modern neural architectures for solving partial differential equations (PDEs) in a data-driven way by minimizing a PDE-based loss, cf.~\citet{raissi2019physics}. % for a general introduction. 
Consider parameterized PDEs of the general form
\begin{equation}
    f(t, {\bs x}|\bs \lambda) :=  \partial_t {\bs u}(t, \bs x) +  P({\bs u}|{\bs \lambda}) = 0,
    \label{eq:PiNN_hardconstraints}
\end{equation}
where $P$ is a non-linear operator parameterized by ${\bs \lambda}$, and $\partial_t$ is the partial time derivative w.r.t. $t \in [0, T]$.
The position ${\bs x} \in \Omega$ is defined on a spatial domain $\Omega \subseteq \mathbb{R}^d$. %, with time $t \in [0, T]$.% are  and a time interval up to limiting time $T$. 
The PDE is subject to initial condition $g_0$
\begin{equation}
    {\bs u}(0, \bs x) = g_0(\bs x)
\end{equation}
for ${\bs x} \in \Omega$, and boundary conditions $g_{\partial\Omega}$
\begin{equation}
    {\bs u}(t, {\bs x}) = g_{\partial\Omega}({\bs x})
\end{equation}
for ${\bs x} \in \partial\Omega$ and $t \in [0, T]$. The main idea of PINNs consists in approximating $\bs u(t, {\bs x})$ (and hence $f(t, \bs x)$) with a neural network given a small set of $N$ noisy observations ${\bs u}_\text{obs}$
\begin{equation}
     {\bs u}(\txi{i}) + \epsilon^{(i)} = {\bs u}_\text{obs}^{(i)}
\end{equation}
with noise $\epsilon^{(i)} \ll \bs u^{(i)}$ $\forall i \in \{0, 1, \dots, N\}$. 
This allows us to consider the following two important problem settings: If $\bs \lambda$  is known, the PDE is fully specified, and we aim to find a solution ${\bs u}$ in a data-driven manner by training a neural network. The PDE takes the role of a regularizer, where the particular physical laws provide our prior information.
A second setting considers the inverse learning of the parameters $\lambda$ by including them into the optimization process in order to infer physical properties such as the viscosity coefficient of a fluid \citep{jagtap2020conservative}.
Initial work on solving time-independent PDEs with neural networks with such PDE-based penalties was pioneered by \citet{dissanayake1994neural} and \citet{van1995neural}, with later adoptions such as \citet{parisi2003solving} extending it to non-steady and time-dependent settings.
% %

\paragraph{Loss functions.}
Typically, PINNs approximate $f(t, {\bs x})$ by the network ${f}_{\Theta}(t, {\bs x})$ in which the parameters $\Theta$ are adjusted by minimizing the combined loss of (i) reconstructing available observations ($L_\text{obs}$), (ii) softly enforcing the PDE constraints on the domain ($L_{f}$), and (iii) fulfilling the boundary ($L_b$) and initial conditions ($L_{\text{init}}$), i.e.
\begin{equation}
\label{eq:PINN_argmin}
    {\Theta} = \underset{\Theta}{\arg\min} \left[ 
          w_1{L}_\text{obs}(\bs X, \bs t, \bs u_\text{obs}, \Theta) 
        + w_2{L}_{f}(\Theta) 
        + w_3{L}_{b}(\Theta)
        + w_4 L_{\text{init}}(\Theta)
    \right], 
\end{equation}
with loss weights $w_i \in \mathbb{R}_{\geq 0}$. 
A common choice for $L_\text{obs}$, $L_b$, and $L_{\text{init}}$ is the expected L$^2$ loss, approximated via the average L$^2$ loss over the observations and via sampled boundary and initial conditions, respectively.
It should be noted that the formulation of the forward and inverse problem are identical in this setting, as observations and initial conditions are implemented in a similar manner.

\paragraph{Enforcing the PDE.}
Although PINNs are by nature mesh-free, %obtaining 
the PDE loss $L_{f}$ in Eq.~\ref{eq:PINN_argmin} used for the soft enforcement of Eq.~\ref{eq:PiNN_hardconstraints} requires a similar discretization step for approximating an integral over the continuous signal domain,
\begin{equation}
    \label{eq:PINN_pde_loss}
    {L}_{f}(\Theta) 
    \!=\!\frac{1}{|[0,T] \times \Omega|} \int \limits_{t=0}^T \!\int \limits_{\Omega} \!\!||{f}_\Theta(t, \bs x)||_2^2 \ \!d\bs x \ dt
    \!=\! E_{p(t, {\bs x})}\Big[\!||{f}_\Theta(t, {\bs x})||_2^2\!\Big]
    \!\approx\! \frac{1}{n} \sum_{i=1}^{n} \!  ||{f}_\Theta(t_i, {\bs x_i})||_2^2
\end{equation} 
with $p(t, \bs x)$ being supported on $[0,T] \times \Omega$.
The points  $\{(\txi{j})\}_{j=1}^n \subset [0,T] \times \Omega$  on which the PDE loss is evaluated are commonly referred to as \textit{collocation points}. 
This formulation of PINNs for solving Eq.~\ref{eq:PiNN_hardconstraints} is an Eulerian one, as the function $f_\Theta$ is updated by evaluating the PDE with respect to collocation points fixed in space.
%Initial approaches for solving differential equations with neural networks relied on a fixed grid for selecting $\{(\txi{j})\}_{j=1}^n$ \citep{lagaris1998artificial, rudd2013solving, lagaris2000neural}. 
% for a better coverage of the full input domain, i.e. by sampling from $p(t,{\bs x}) \propto 1$.
%The work of \citet{nabian2021efficient} explores Importance Sampling based on Inverse Transform sampling for a more sample efficient evaluation of Eq.~\ref{eq:PINN_pde_loss}.
Initial approaches for selecting the collocation points in PINNs relied on a fixed grid %for selecting collocation points
\citep{lagaris1998artificial, rudd2013solving, lagaris2000neural}, followed up by work proposing stochastic estimates of the integral via (Quasi-) Monte Carlo methods \citep{sirignano2018dgm, lu2021deepxde, chen2019quasi} or Latin Hypercube sampling \citep{raissi2019physics}.
However, these approaches to Eulerian PINNs cannot be directly applied if there are no known boundaries or boundary conditions, e.g. for $\Omega = \mathbb{R}^d$.
Additionally, problems can arise if the constrained region is large compared to the area of interest.
Considering for example the shock wave (of a compressible gas) in a comparably large space, most collocation points would fall into areas of low density. %outside of relevant regions.
We argue that due to the locality of particle interactions, the regions with higher density are more relevant for regularizing the network.
%While manual selection of the collocation points could be possible in some settings, it usually requires some a-priori knowledge about the solution. 

To address these shortcomings of previous methods, we propose a mesh-free and adaptive approach for sampling collocation points, illustrated on the example of compressible fluids. 
By changing $p(t, \bs x)$ to the distribution over the particle positions in the fluid we effectively change the loss functional in Eq.~\ref{eq:PINN_pde_loss}.
We then generalize to other settings, such as thermodynamics, by interpreting a positive, scalar quantity of interest with a finite integral as a particle density.
Within this work we specifically focus on PDEs that can be derived based on local particle interactions or 
can be shown to be equivalent to such a view, as for example is the case for the heat equation with its connection to particle diffusion.
Notably, we do not require the introduction of Lagrangian updates, as classical mesh-free methods do,
which would be based on evaluating the PDE with respect to moving particles (see also section \ref{sec:related_work}).
%which would evaluate the PDE by considering changes of physical quantities with respect to moving particles,
%and 

\paragraph{Main contributions.}
The main contributions of this paper are as follows:
\begin{itemize}
    %\item We demonstrate failure modes of PINNs with uniform sampling strategies in settings with spatially sparse signals and unbounded signal domains.
    \item We demonstrate that PINNs with uniform sampling strategies (and refinement methods based on uniform proposals) fail in settings with spatially sparse signals as well as in unbounded signal domains; these problems can severely degrade the network's predictive performance.
    \item 
    In order to overcome these limitations of existing approaches, we propose a truly mesh-free version of Eulerian PINNs, in which the collocation points are sampled using physics-motivated MCMC methods.
    By staying within the Eulerian framework, we avoid conceptual challenges of classical mesh-free methods based on Lagrangian updates such as the enforcement of boundary conditions.
   %, as well as distributions of Brownian particles with drift.
    %\item
    %Different to classical mesh-free methods, our proposal is constructed in the Eulerian formulation of the problem and does not rely on Lagrangian updates, circumventing conceptual challenges in Lagrangian methods such as the enforcement of boundary conditions.
    \item
    The proposed model is applicable to a huge range of dynamical systems governed by PDEs that share an underlying microscopic particle description, such as several hydrodynamic, electro- and thermo-dynamic problems.
    \item 
    We rigorously evaluate and compare our proposed method with existing approaches in high-dimensional settings.
    Compared to existing mesh refinement methods, significantly fewer collocation points are required to achieve similar or better predictive performances, while still being more flexible.
\end{itemize}

\section{Related Work}
\label{sec:related_work}
% \paragraph{Adaptive Mesh Refinement.}
% Classical numerical PDE methods 
% require a discretization process 
%  based on a computational mesh or grid.
% The resolution of the mesh sets a fundamental trade-off between computational (and memory) 
% complexity and %the 
% accuracy. % of the model.
% In practice, most computational effort is spent close to boundary regions. %in areas of interaction with difficult boundaries.
% But even in settings without complex boundary geometries, local refinements and hierarchical meshes can be preferable over uniform Cartesian grids \citep{hirsch2007numerical, plewa2005adaptive, pons2019adaptive}.
% Furthermore, a wide range of adaptive mesh refinement methods has been proposed, that rely on heuristics to define regions of interest for the mesh refinement \citep{plewa2005adaptive, pons2019adaptive}. 

\paragraph{Mesh-Free Fluid Dynamics.}
Classical mesh-free approaches in computational 
fluid dynamics are based on non-parametric function representations, with Smoothed Particle Hydrodynamics (SPH) \citep{lind2020review, gingold1977smoothed} being the most prominent example.
In SPH, %the generation of a mesh such as in Finite Volume or Finite Difference methods is avoided by 
fluid properties such as the density and pressure are represented by a discrete set of particles and interpolated using a smoothing kernel function.
For updating the function forward in time, the particles have to be propagated according to the Lagrangian formulation of the PDE, relying on the kernel for computing spatial derivatives.
One of the benefits of such a representation is that mass is conserved by construction. 
However, Lagrangian updates become challenging when enforcing boundary conditions, requiring the introduction of ad-hoc "dummy" or "mirror" particles \citep{lind2020review}.
Instead, we present a mesh-free, particle-based, PINN that does not require Lagrangian updates, and is already applicable in the Eulerian formulation.
It should be noted that the proposed pdPINNs can in principle be combined with Lagrangian updates such as proposed by \citet{raissi2019physics} and later by \citet{wessels2020neural}.
%Although PINNs were initially based on the Eulerian formulation, similar Lagrangian updates could be included, as showcased by the Runge Kutta method presented in \citet{raissi2019physics} and later extended by \citet{wessels2020neural}.%, but are not necessarily required for updating the function, i.e. the network.
%An example for such an approach is the Runge Kutta method presented in \citet{raissi2019physics} and later extended by \citet{wessels2020neural}, which could in principle be combined with our proposed method for (re-)sampling the particle positions.
%Our approach of sampling the particles can be combined with either the Eulerian or Lagrangian formulation and is agnostic in that regard
%As the intention of this work is to improve upon current Eulerian PINNs, we consider the comparison and extension to the Lagrangian formalism to be out of scope and refer for this to future work.
But as the intention of this work is to improve upon current Eulerian PINNs, we refer to future work for the comparison and extension to the Lagrangian formalism.
%But as this work does not intend to provide a comparison between Eulerian and Lagrangian approaches in PINNs, we will focus on Eulerian formulations of the PDEs due to their ease of implementation. 
%Due to the particle representation, SPH conserve mass by construction. Major limitations however include difficulties in constraining the functions, e.g. for enforcing boundary conditions, with current solutions often requiring the cumbersome introduction of dummy or mirror particles.

\paragraph{Alternative Meshes and Losses for PINNs.}
Recent work proposes local refinement methods for PINNs %during the training process 
by adding more samples within regions of high error \citep{lu2021deepxde, tadiparthi2021optimal}.
Residual adaptive refinement (RAR) is suggested by \citet{lu2021deepxde}, which is based on regularly evaluating the PDE loss on a set of uniformly drawn samples.
The locations corresponding to the highest PDE loss are then added to the set of collocation points used in training.
\citet[preprint]{tadiparthi2021optimal} further enhance RAR by learning a linear map between the uniform distribution and the distribution over the PDE loss by optimizing an optimal transport objective.
By sampling uniformly and subsequently transforming these samples, it is attempted to focus on regions of higher error. Due to the conceptual similarity to RAR, we will denote this method as "OT-RAR".
The work of \citet{nabian2021efficient} explores Importance Sampling based on the (unnormalized) proposal distribution $||f_\Theta(t, \boldsymbol{x})||_2^2$ for a more sample efficient evaluation of Eq.~\ref{eq:PINN_pde_loss}. Samples are drawn using a variation of Inverse Transform sampling \citep{steele1987non}.

However, in all these cases the underlying mechanism for exploring regions of high error is based on (quasi-) uniform sampling within the boundaries.
%Furthermore, the distribution of the (absolute or squared) PDE-loss can often be highly complex and may require the evaluation of higher order derivatives.
As such, they do not resolve the issues of unknown boundaries and will furthermore be infeasible in higher dimensions. 

\paragraph{Kinetic Theory: From particles to PDEs.}
Kinetic theory shows that essential conservation laws of fluids can be derived from a microscopic (or molecular) viewpoint \citep{born1946general}. 
%That is, the 
Interactions describing the dynamics of a fluid are described starting from a set of individual particles.
The basis of this approach is the so-called
\textit{molecular distribution function} $\molecDist$ over phase space, i.e. $\molecDist(t, \bs{x}, \bs{v})$ such that 
\begin{equation}
    \int_{\Delta \boldsymbol{x}} \int_{\Delta \bs{v}} \molecDist(t, \bs{x}, \bs{v}) d\bs{v} d\bs{x}    
\end{equation}
is the probability that a molecule with a velocity within $\Delta v = \Delta v_1 \Delta v_2 \Delta v_3$ occupies the volume $\Delta \bs{x} = \Delta x_1 \Delta x_2 \Delta x_3$.
Based on this distribution function, it is possible to define common quantities as the (mass or particle) density, (local mean) velocity, and macroscopic PDEs by considering the local interactions of individual particles.
The one-particle phase space is commonly known from its application in the Boltzmann equation for modelling two-body interactions describing gases \citep{green1956boltzmann} and active matter  (e.g. flocks of birds) \citep{bertin2006boltzmann}.
The more general form including higher interaction terms is necessary for deriving conservation laws of liquids \citep{born1946general}.

\section{Particle-density PINNs}
\label{sec:methodology}
In this section we introduce the concept of mesh-free \textit{particle-density PINNs (pdPINNs)}. 
Firstly, we examine limitations of the common PDE loss in Eq.~\ref{eq:PINN_pde_loss}
and, secondly, we present a solution by integrating over the position of particles instead of the full support of the signal domain.

% LMC -> 
%The presented approach is applicable to a wide range 
%of advection-diffusion problems that are crucial in many hydrodynamic flow problems 
%as well as electro- and thermodynamic ones and beyond, as for instance in applications of the Fokker-Plank equation.
The underlying assumption of our approach is that the dynamics described by the PDE can be explained in terms of local interactions of particles.
This is the case, for instance, for commonly considered dynamics of gases, liquids or active particles \citep{hoover2003links, toner1995long}.
%LMC Furthermore, the naive solution of selecting a sufficiently large area with $a\ll b$ for resolving (2.) would amplify the sparsity in (1.).

%LMC Firstly, we examine limitations of the common PDE loss in Eq.~\ref{eq:PINN_pde_loss}
%LMC and, secondly, we present a solution by integrating over the position of particles instead of the full support of the signal domain.

%LMC\subsection{Modeling Compressible Fluids}
%LMC\label{sec:PINN}
\paragraph{Existing limitations of Eulerian PINNs.}
Consider the problem of modeling a (possibly non-steady) compressible fluid, i.e. a fluid with a spatially and temporally evolving density $\rho(t, \bs x)$ and velocity $\bs v(t, \bs x)$. 
For the sake of notational brevity, we will denote these by $\rho$ and $\bs v$. % in the following.
%\textcolor{red}{Given noisy observations, our particular interest lies in the prediction of particle movement, hence in modelling the density $\rho$ and potentially other quantities such as the velocity or pressure.}
Given noisy observations, our particular interest lies in the prediction of particle movements, hence in the approximation of the density (and potentially other physical quantities) with a neural network $\rho_\Theta$. Additional quantities such as the velocity or pressure might also be observed and modeled.
%LMC and thus density $\rho$.
%LMC Given noisy observations of the density $\rho^{(i)}$, and potentially other quantities such as the velocity or pressure, we can train a Neural Network $\rho_\Theta$ to approximate it.
%we want to learn the true density $\rho$. % via the neural network $\rho_\Theta$.

\begin{figure}
  \centering
    \begin{subfigure}{.4\textwidth}
      \centering
        \includegraphics[width=0.9\linewidth]{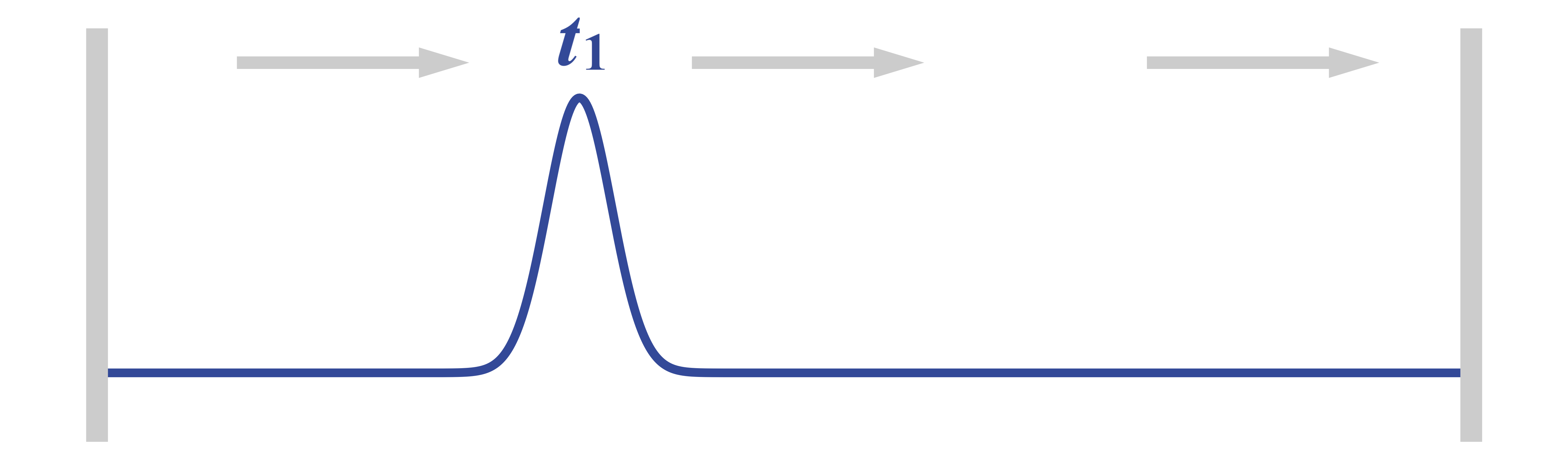}
      \caption{}
      \label{fig:advection_gt_t0}
    \end{subfigure}%
    \begin{subfigure}{.4\textwidth}
      \centering
        \includegraphics[width=0.9\linewidth]{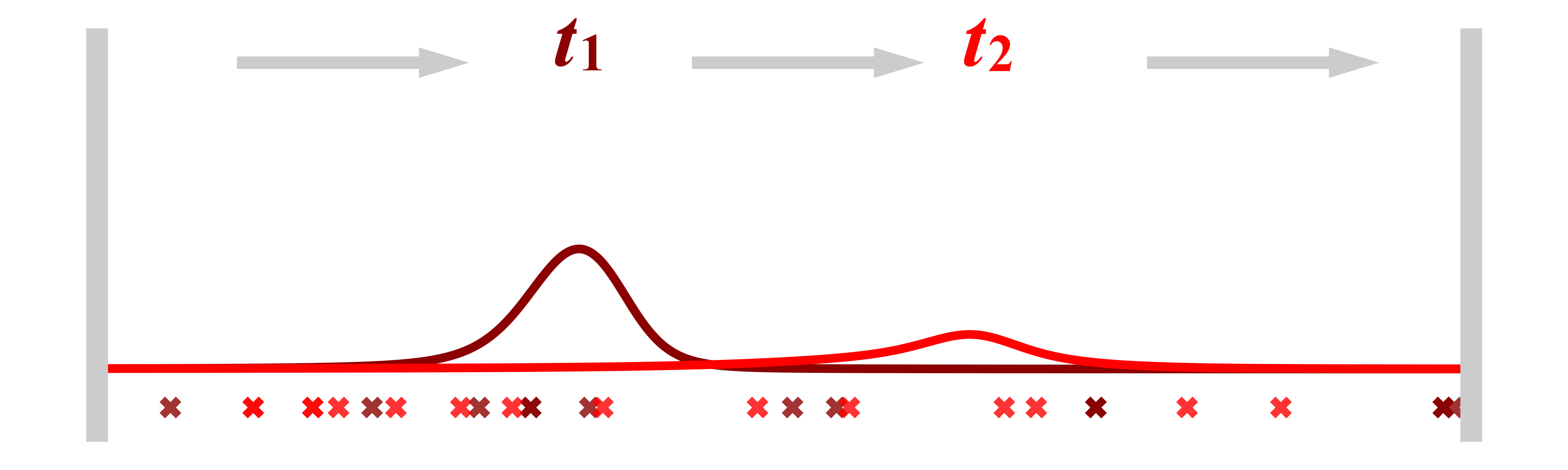}
      \caption{}
      \label{fig:advection_unif}
    \end{subfigure}
    \begin{subfigure}{.4\textwidth}
      \centering
        \includegraphics[width=0.9\linewidth]{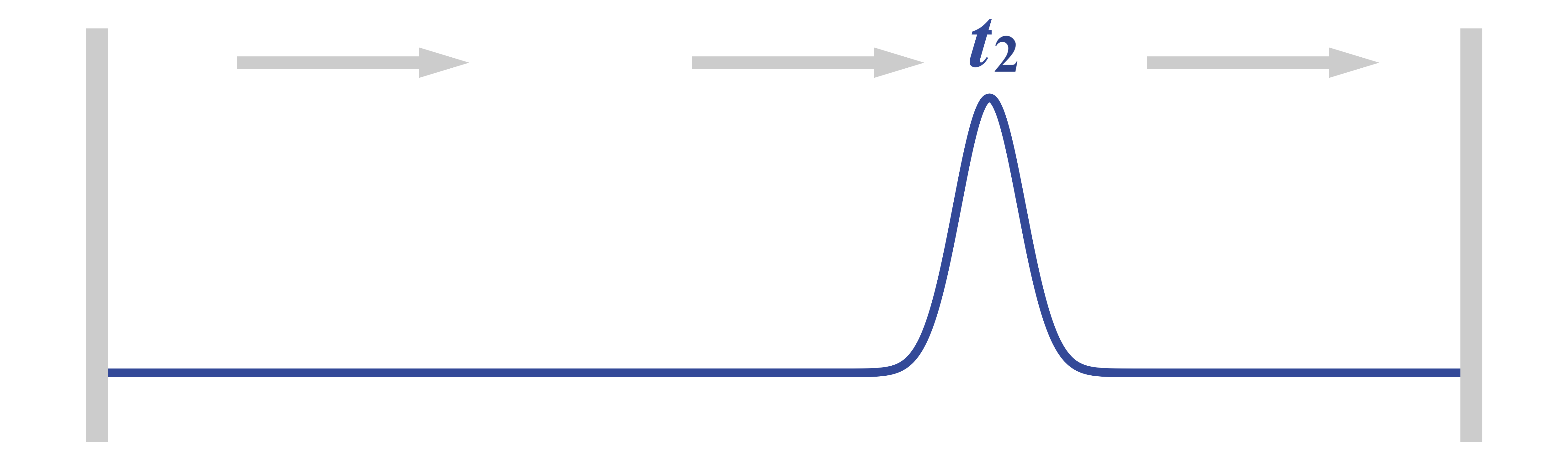}
      \caption{}
      \label{fig:advection_gt_T}
    \end{subfigure}%
    \begin{subfigure}{.4\textwidth}
      \centering
        \includegraphics[width=0.9\linewidth]{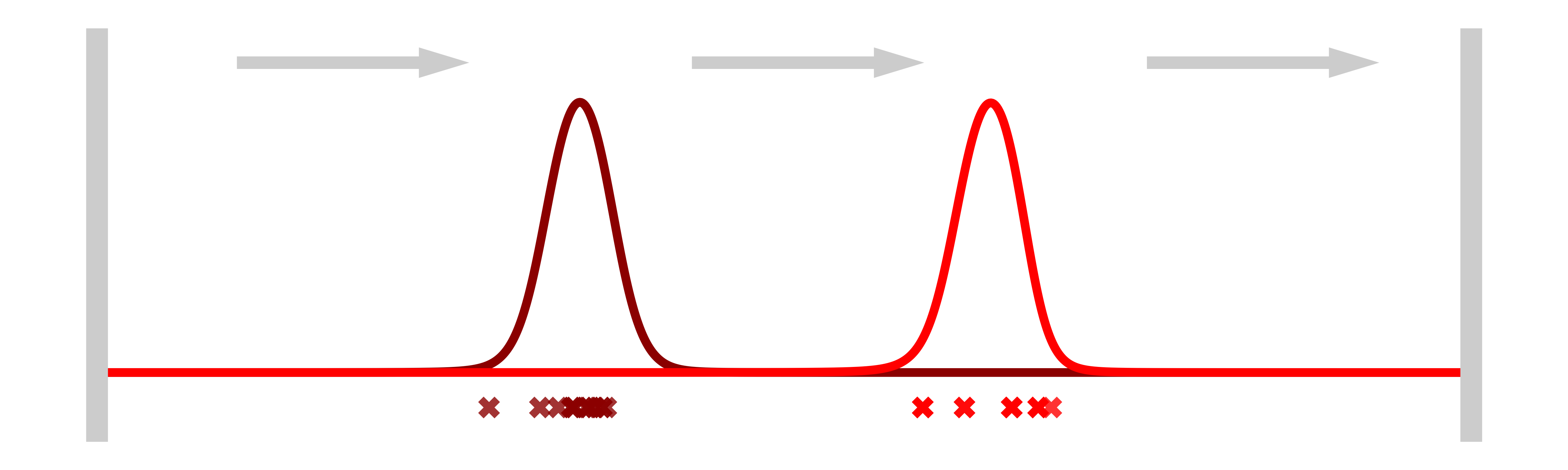}
      \caption{}
      \label{fig:advection_mcmc}
    \end{subfigure}
  \caption{Advection experiment in 1D: (a) ground truth at time $t_1$ and (c) time $t_2$, (b) density prediction with uniform collocation points and (d) particle-density-based collocation points for $t\in\{t_1, t_2\}$, with crosses indicating sampled points. %Cross points represent sampled collocation points at $t_1$ and $t_2$ with two different red shades.
  }
 \label{fig:advection}
\end{figure}

% Maxim's suggestion:
Commonly, the PDE then serves as a physics-based regularizer of the network by enforcing the PDE loss $L_f$ in Eq.~\ref{eq:PINN_pde_loss} during standard PINN training. 
For this, $L_f$ is evaluated on a set of collocation points that are, for example, uniformly distributed on a bounded region.
However, the limitations of this approach already become apparent when considering a simple advection problem defined by the following PDE:
\begin{equation}
    \label{eq:advection}
    \partial_t \rho +   \bs v\cdot (\nabla\rho)= 0.
\end{equation}
Figure \ref{fig:advection} illustrates a one-dimensional case on the domain $[0,T]\times \Omega$, with $\Omega = \mathbb{R}$, and a known constant velocity $v\propto1$.
% Spatially fixed sensors measure 
We measure the density $\rho^{(i)}$ at different (spatially fixed) points in time and space $\{(t^{(i)}, \bs x^{(i)})\}$, on which a neural network $\rho_{\Theta}(t, \bs x)$ is trained.
For optimizing the standard PDE loss $L_f$ as given in Eq.~\ref{eq:PINN_pde_loss}, we would require a bounded region $\Omega_{\mathcal{B}}:=[a,b] \subset \Omega$ with $a<b$ and $a,b \in \mathbb{R}$. 
This, in turn, leads to two issues:
\begin{enumerate}%[i]
    \item 
    %the moving density occupies a small subset of $\Omega$
    %Our region of interest, 
    Since the moving density occupies a small subset of $\Omega$,
    uniformly distributed collocation points within $\Omega_{\mathcal{B}}$ will enforce Eq.~\ref{eq:advection} in areas with low-density.
    This results in insufficient regularization of $\rho_\Theta$.
    %In general, the moving density occupies a small subset of $\Omega$ and enforcing Eq.~\ref{eq:advection} in areas with low-density would have little, if any, regularizing effect on $\rho_\Theta$}.
    \item Defining a suitable bounded region $\Omega_{\mathcal{B}}$ requires a priori knowledge about the solution of the PDE, % time frame, velocity and solution of the underlying PDE, all of 
    which is generally not available. 
    Choosing too tight boundaries would lead to large parts of the density moving out of the considered area $\Omega_{\mathcal{B}}$.
    Too large boundaries would instead lead to poor regularization as this would worsen the sparsity problem in issue (1.).
    %  Too large boundaries would instead lead to poor regularization as highlighted in issue (1.).
    %Otherwise, the main bulk of density might move out of the considered area $\Omega_{\mathcal{B}}$.
\end{enumerate}
%LMC\begin{enumerate}
%LMC    \item Our region of interest, the moving density, is concentrated on a small subset of $\Omega$.
%LMC    Enforcing Eq.~\ref{eq:advection} in areas with low-density has little regularizing effect on $\rho_\Theta$.
%LMC    \item 
%LMC    In order to define appropriate boundaries of $\Omega_{\mathcal{B}}$, we require prior knowledge about the solution which depends on the time frame, velocity, and the underlying PDE. 
%LMC    Otherwise, the main bulk of density might move out of the considered area $\Omega_{\mathcal{B}}$.
%LMC\end{enumerate}
In practice, most Eulerian PINNs approaches opt for naively defining a sufficiently wide region $\Omega_{\mathcal{B}}$, resulting in a poor reconstruction.
%, which however has the undesired effect outlined in (i), often resulting in an insufficiently regularized network.
In the context of our advection problem, this is showcased in Figure \ref{fig:advection_unif}. % where the density network is barely regularized, leading to poor reconstruction.
To properly resolve the aforementioned issues, one should (i) focus on areas that have a relevant regularizing effect on the prediction of $\rho_\Theta$ and (ii) adapt to the fluid movements without being restricted to a predefined mesh.  
\vfill
%\textcolor{red}{
%We propose to directly draw samples from the particle density, which completely removes the need of defining ad-hoc boundaries while flexibly allowing to focus on highly relevant regions, i.e. those with high density.}
%We propose to sample from the occupation probability of particles, namely the 
%This completely removes the need of defining ad-hoc boundaries while flexibly allowing to focus on highly relevant regions, i.e. those that are more densely populated. %with higher occupation probability.
%Then the problem is reduced to minimize the expectation over the occupation probability of particles over the whole time frame:
%LMC We note that for issue (1.), the dynamics of fluids are assumed to be mainly caused by local interactions of particles, which is the case for commonly considered dynamics of gases, liquids or active particles \citep{hoover2003links, toner1995long}.
%LMC Furthermore, the naive solution of selecting a sufficiently large area with $a\ll b$ for resolving (2.) would amplify the sparsity in (1.).
%
%
%
%LMC Therefore,  proper resolution to these issues should be able to (i) focus on areas that have a relevant regularizing effect on the prediction of $\rho_\Theta$ and (ii) adapt to the fluid movements without being restricted to a predefined mesh.  
%LMC Consider the hypothetical setting of knowing %Let us assume that we had access to
%LMC the positions of each of the $N$ total particles via the function $\bs x_j(t)$, with $j\in \{0,1,\dots,N\}$.
%LMC We would then propose to use the following loss over these positions within the considered time frame $[0, T]$:
%
%LMC\begin{equation}
%LMC\label{eq:L_PD_sum}
%LMC     {L}_{\text{pd}}(\Theta)=\frac{1}{T} \int_{t=0}^{T} \frac{1}{N}\sum_{j=0}^{N} \Big|\Big|f_\Theta\Big(t, \bs x_{j}(t)\Big)\Big|\Big|_2^2 \ \ dt.
%LMC\end{equation}
%
%LMC It can be seen that the average in Eq.~\ref{eq:L_PD_sum} is a sample-based estimator of an expected PDE loss.
%
%LMC Namely, the expectation over the occupation probability of particles in the whole time frame, denoted by the molecular distribution $\molecDist(t, \bs x) = \molecDist(t) \molecDist(\bs x| t)$ introduced in section \ref{sec:related_work}:
\paragraph{Mesh-Free Eulerian PINNs.} We thus propose to reformulate the PDE loss in Eq.~\ref{eq:PINN_pde_loss} as the expectation of $||{f}_\Theta(t, \bs x)||_2^2$ with respect to the
molecular distribution $\molecDist(t, \bs x)$ introduced in the related work section \ref{sec:related_work}:
\begin{equation}
    %L_{\text{pd}}(\Theta) \approx \int_{t=0}^T \molecDist(t) \int_{\Omega}  \molecDist(\bs x | t)\Big[||f_\Theta(t, \bs x_{j}(t))||_2^2 \Big] \ \ dt,
    L_{\text{pd}}(\Theta) \approx \int_{t=0}^T  \int_{\Omega}  \molecDist(t, \bs x)\Big[||f_\Theta(t, \bs x)||_2^2 \Big] \ d\bs x \ dt.
\end{equation}
This completely removes the need of defining ad-hoc boundaries while providing the ability to flexibly focus on highly relevant regions, i.e. those that are more densely populated.
As the particle density corresponds directly to the occupation probability of a molecule $\molecDist(t, \bs x)$ with a changed normalization constant,
we can estimate $L_\text{pd}$ via samples drawn from the normalized particle density, which is denoted as $\rho_N$.
For homogeneous fluids, this coincides with the normalized mass density. % when assuming a homogeneous fluid, both of which we will .
%LMC \subsection{Particle-Density PINNs}
%LMC \label{subsec:pdPINNs}

In summary, we propose to draw collocation points from the normalized density:
%
%\vspace{-0.5mm}
\begin{equation}
    (t_i, \bs{x}_i)  \sim  \rho_N(t, \bs{x}) = 
    \tfrac{1}{Z} 
    % 1/Z 
    \rho(t, \bs x).
\end{equation}
The true particle positions and the density $\rho_N$ are however unknown in practice. % everywhere
Instead, we have to rely on the learned density ${\rho}_{\Theta}(t, {\bs x})$ as a proxy provided by the neural network. 
We denote the associated normalized PDF by $q_{\Theta}(t, {\bs x}) = \frac{1}{Z'} {\rho}_{\Theta}(t, {\bs x})$ with support on $[0, T] \times \Omega$. 
The PDE loss is then defined as the expectation w.r.t.~$q_{\Theta}(t, {\bs x})$: 
\begin{equation}
    \label{eq:PINN_pde_loss_density}
    {L}_{pd}(\Theta) = \mathbb{E}_{q_{\Theta}(t, \bs x)}\Big[ ||{f}_\Theta(t, \bs x)||_2^2 \Big] 
   % = \int_{t=0}^T q_{\Theta}(t) \int_{\Omega}  q_{\Theta}({\bs x | t})\ ||{f}_\Theta(\bs x,t)||_2^2 \ d\bs x \ dt. 
    = \int_{t=0}^T \int_{\Omega}  q_{\Theta}({t, \bs x})\ ||{f}_\Theta(\bs x,t)||_2^2 \ d\bs x \ dt. 
\end{equation}
In order to approximate this integral, samples need to be drawn from  $q_\Theta(t, \bs x)$. 
This can be done in a principled way by using dynamic Monte Carlo methods, despite the fact that the normalization constant $Z$ is unknown.
%we will explore the use of advanced Monte Carlo methods that include derivative information,
%such as Hamiltonian Monte Carlo \citep{neal2011mcmc}.
%This seamlessly integrates with the PINN framework, as the derivatives of $\rho_\Theta(t, {\bs x})$ with respect to the inputs are readily available via automatic differentiation.
%Such a sampling procedure additionally eliminates the need for $\Omega$ to be a bounded domain in settings where such boundaries are not available.
We highlight that, in contrast to the mesh-based loss in Eq.~\ref{eq:PINN_pde_loss}, the loss in Eq.~\ref{eq:PINN_pde_loss_density} is also suitable for problems on unbounded domains such as $\Omega = \mathbb{R}^d$.
% ,and potentially arbitrary boundaries have to be introduced. 

 %, both compressible fluid dynamics settings and more general settings 
%as we elaborate on below.

\paragraph{Applicability of pdPINNs.}
Although motivated in the context of an advection problem, the proposed approach is generally applicable to a wide range of PDEs.
The advection equation \ref{eq:advection} can be seen as a special case of mass conservation (assuming $\nabla \cdot \bs v = 0 $), which is one of the fundamental physical principles expressed as a \textit{continuity equation}.
This continuity equation relates temporal changes of the fluid density $\rho$ to spatial changes of the flux density $\rho \bs v$ through
\begin{equation}
    \label{eq:continuity_equation_masscons}
    \partial_t \rho +  \nabla \cdot (\rho \bs v)=  0.
\end{equation}
%The presented approach is applicable to a wide range of 
%Advection-diffusion problems that are crucial in many hydrodynamic flow problems as well as electro- and thermodynamic ones and beyond, as for instance in applications of the Fokker-Plank equation.
%Another common physical process that is suited for our approach is diffusion, such as in the Heat Equation, where local interactions of particles give rise to the PDE (as established by Fick's second law):
Another common physical process that is suited for our approach is diffusion, such as in the Heat Equation, where local interactions of particles give rise to the following PDE (as established by Fick's second law):
%The diffusion of the temperature $T$ within a homogeneous medium is described by a parabolic PDE defined as
%The heat equation is a parabolic PDE defined as
\begin{equation}
\label{eq:heat_equation}
\partial_t T - \alpha \nabla^2 T = 0,
\end{equation}
where $T$ denotes the temperature interpreted as density, $\alpha$ the thermal (or mass) diffusivity, and $\nabla^2$ the Laplacian operator.
By introducing additional constraints to the diffusion and mass-conservation, one can describe viscous fluids with the Navier-Stokes equations or even self-propelled, active particles, for which Toner and Tu \citep{toner1995long,tu1998sound,toner1998flocks} introduced hydrodynamic equations.
Other possible applications involve Maxwell's equations for conservation of charge in electrodynamics,
as well as the distribution of Brownian particles with drift described by the Fokker-Planck equations. 
In general, our method is applicable in settings where (i) a non-negative scalar field (with a finite integral) of interest can be interpreted as a particle density,
and (ii) the local interactions of these particles give rise to the considered PDEs.

\section{Model and Implementation}
\label{sec:model_and_implementation}
A wide range of different network architectures and optimization strategies for PINNs have emerged. % \citep{sitzmann2020implicit, tancik2020fourier, wang2021eigenvector, wang2021understanding}.
They emphasize well-behaved derivatives with respect to the input domain \citep{sitzmann2020implicit}, allow higher expressivity for modelling high frequency data \citep{tancik2020fourier, wang2021eigenvector}, or resolve gradient pathologies within PINNs \citep{wang2021understanding}.
As our method does not rely on a specific architecture, any such improvement can be easily combined with the proposed pdPINNs.
For the experiments in this submission we will use simple fully-connected networks with sinusoidal \citep{sitzmann2020implicit} or tanh activations (see section \ref{sec:experiments}).
%Although we use simple fully-connected networks with sinusoidal or tanh activations (see section \ref{sec:experiments}), our method can be freely combined with any architectural improvements.
%In contrast, our proposed approach is independent of the particular choice of the architecture and we thus use simple fully-connected networks with sinusoidal or tanh activations (see section \ref{sec:experiments}).
%In contrast, our proposed approach is independent of the particular choice of the architecture and we thus use simple fully-connected networks with sinusoidal or tanh activations (see section \ref{sec:experiments}).
%the particular choice of the network architecture is orthogonal to our proposed method. % and should be compatible with any further improvements. 
%For the experiments in section \ref{sec:experiments},
% Monika
%in the next section, 
%we mostly choose fully-connected networks with either sinusoidal or tanh activations. 

\paragraph{Finite total density.}
For reformulating the predicted density $\rho_\Theta$ as a probability, we have to ensure non-negativity as well as a finite integral over the input domain $\Omega$.
Non-negativity can for example be achieved via a squared activation function after the last layer. % (or alternatively with an exponential or softplus activation). 
An additional bounded activation function $g$ is then added, which guarantees the output to be within a pre-specified range $[0, c_{max}]$.
%By further multiplying the bounded output of the network with a Gaussian kernel, we enforce the integral over $\mathbb{R}^d$ to be finite.
The integral $\mathbb{R}^d$ can then be enforced to be finite by multiplying the bounded output with a Gaussian kernel.
Summarizing these three steps, let $\tilde\rho_\Theta$ denote the output of the last layer of our fully connected neural
network and $p_{\text{gauss}}(\bs x)=\mathcal{N}(\bs x;\mu, \Sigma)$, then we predict the density $\rho_\Theta$ as
\begin{equation}
    \rho_\Theta(t, \bs x) =  p_{\text{gauss}}(\bs x)\  g(\tilde{\rho}_\Theta(t, \bs x)^2) \leq c_{\text{max}}p_{\text{gauss}}(\bs x).
\end{equation}
%
%In practice a large value for $c_max$ is used, which did not affect any of our experiments.
% In practice a sufficiently large $c_{max}$ doesn't have any effect on the model
In practice, the choice of $c_{\text{max}}$ does not affect the model as long as it is sufficiently large.
The used mean $\mu$ and covariance $\Sigma$ are maximum likelihood estimates based on the observations $\bs x$, i.e. the sample mean $\bar{\bs x}$ and covariance $\bar{ \Sigma}$ of the sensor locations.
%In this way we enforce a zero-prior in regions far away from the observed data.
To allow more flexibility 
%avoid being too restrictive 
in the network, we add a scaled identity matrix to the covariance $\Sigma = \bar{ \Sigma} + c \cdot I$, %which may be manually adjusted depending on the problem at hand, and 
which can be set to a large value for solving PDEs when only initial conditions, but no observations, are available.

\paragraph{Markov chain Monte Carlo (MCMC) sampling.}
Finally, MCMC methods allow us to draw samples from the unnormalized density $\rho_\Theta(t, \bs x)$.
%We consider several MCMC samplers and emphasize that the choice of a sampler depends on the problem at hand,
%often offering a trade-off between sampling-speed and convergence of the chains.
We consider several MCMC samplers and emphasize that the wide range of well-established methods offer the ability to use a specialized sampler for the considered problem, if the need may arise.
% Finally, we employ MCMC methods to draw samples from the unnormalized density $\rho_\Theta(t, \bs x)$.
% Our approach allows to freely specify the particular sampler according to the problem under study, allowing the user to choose from a wide range of well-established MCMC methods in the literature.
%We consider several MCMC samplers and emphasize that the wide range of well-established methods offer the ability to use a specialized sampler for the considered problem, if the need may arise.
%Although simple sampling strategies might work well for low-dimensional uni-modal densities, more sophisticated samplers are required in high dimensional settings and/or multi-modal settings.
%For the sake of simplicity, we used a random walk Metropolis-Hastings sampler, 
Gradient-based samplers such as %the Metropolis-adjusted Langevin algorithm \citep{xifara2014langevin} or 
Hamiltonian Monte Carlo \citep{duane1987hybrid, betancourt2017conceptual} are particularly suited for our setting, as the gradients of $\rho_{\Theta}$ with respect to the input space are readily available.
For problems where boundaries are known and we have to sample from a constrained region, a bijective transformation is used so that the Markov chain may operate in an unconstrained space \citep{parno2018transport}.
In our experience, both Metropolis Hastings and Hamiltonian Monte Carlo already worked sufficiently well for a wide range of PDEs without requiring much fine-tuning.
We highlight that pdPINNs do not directly depend on MCMC as a sampler, and alternative sampling methods such as modern variational inference schemes \citep{rezende2015variational} can also be directly used as a substitute.
%Alternative sampling methods such as modern variational inference schemes \citep{rezende2015variational} could be attractive alternatives as quick sampling mechanisms.
%, despite providing only an approximation of the target distribution.

%Although the predicted density $\rho_\Theta(t, \bs x)$ approximates $\rho(t, \bs x)$ once trained, it is purely random in its initial, untrained stage.
%As such, we start training the pdPINNs with a warm-up phase in which collocation points are obtained from a pre-specified background distribution of choice, slowly increasing the fraction of samples obtain from $\rho_\Theta(t, \bs x)$.
For details regarding the samplers used and implementation we refer to the Experiments section \ref{sec:experiments} and Appendix section \ref{appendix:background}.

\section{Experiments}
\label{sec:experiments}
In this section we demonstrate the advantages of pdPINNs compared to \textit{uniform sampling}, \textit{importance sampling} \citep{nabian2021efficient} as well as the adaptive refinement methods \textit{RAR} \citep{lu2021deepxde} and \textit{OT-RAR} \citep{tadiparthi2021optimal}.
Despite the term \textit{uniform} sampling, we rely in all our experiments on quasi-random Sobol sequences for more stable behavior in the low samples regime.
%We considered slight variations of the proposed implementations of RAR and OT-RAR to allow a fair comparison with a limited number of admissible samples.
%Specifically, instead of adding to an ever increasing set of collocation points, we limit them to a specified budget.
To guarantee a fair comparison, we considered slight variations of the proposed implementations of RAR and OT-RAR, so that only a limited number of collocation points are used.
For the pdPINNs we consider multiple MCMC schemes, including inverse transform sampling (IT-pdPINN), Metropolis-Hastings (MH-pdPINN), and  Hamiltonian Monte Carlo (HMC-pdPINN) methods.

The models in sections \ref{subsec:exp_compressible_fluid} and \ref{subsec:exp_heat_eq} are implemented in PyTorch \citep{pytorch}, with a custom Python implementation of the MH and Inverse Transform samplers.
For the Fokker-Planck experiment in section \ref{subsec:exp_fp}, we make use of the efficient MCMC implementations provided by TensorFlow probability \citep{abadi2016tensorflow, lao2020tfp} and the utilities of the DeepXDE library \citep{lu2021deepxde}.
%\textcolor{red}{This experiment is implemented in Tensorflow, leveraging the efficient MCMC implementations provided by Tensorflow probability \citep{abadi2016tensorflow, lao2020tfp} and the utilities of the DeepXDE library \citep{lu2021deepxde}.
%The model is implemented in PyTorch \citep{pytorch}, with a custom Python implementation of the MH and Inverse Transform samplers.
%}
%
%
%
More details, as well as further experiments comparing the wall-time of the various samplers, are provided in the Appendix with the code being provided in the supplementary material.
%

%
%\subsection{Particle Simulation of a Compressible Fluid}
\subsection{Mass conservation for simulated particles}
\label{subsec:exp_compressible_fluid}
\begin{figure}[ht!]
    \centering
    \begin{subfigure}{.43\textwidth}
      \flushleft
        \includegraphics[width=\linewidth]{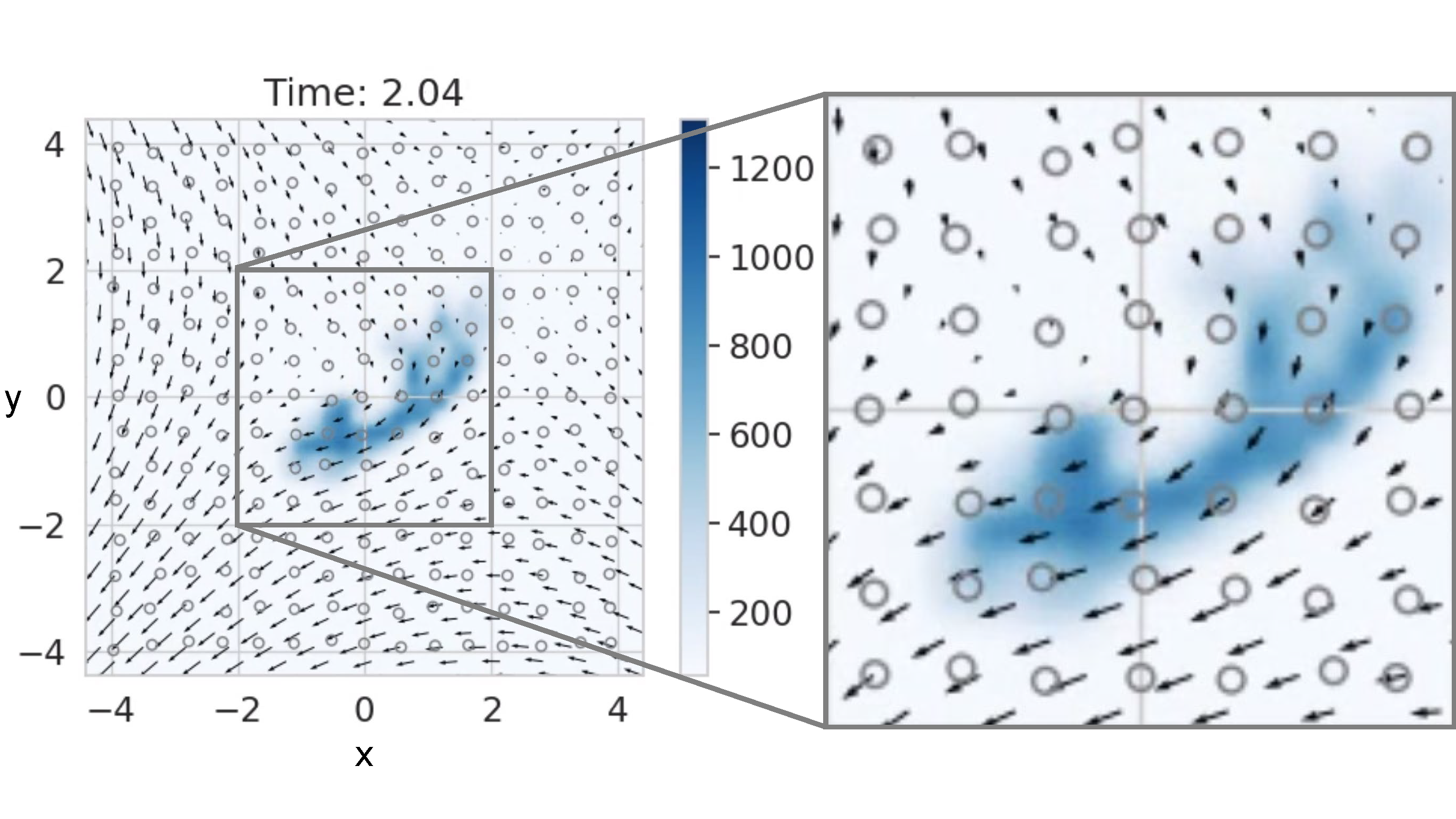}
      \caption{}
      \label{fig:exp3d_birdeye}
    \end{subfigure}%
    \hfill
    \begin{subfigure}{.55\textwidth}
      \flushright
        \includegraphics[width=\linewidth]{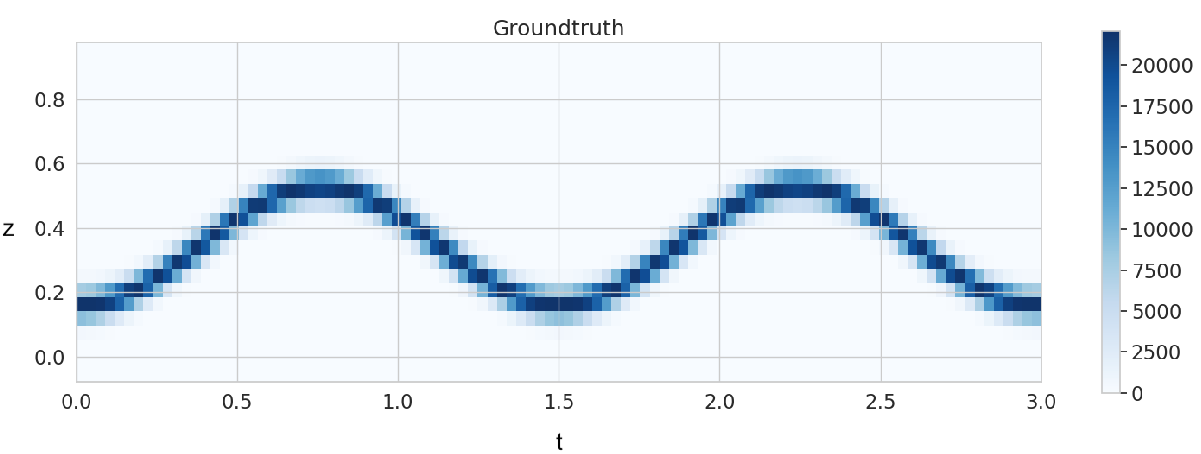}
       \caption{}
       \label{fig:exp3d_zaxis}
    \end{subfigure}%
    \caption{Visualization of the 2D compressible fluid experiment. (a) Bird-eye view of the ground truth particle density. (b) $z$-projection of the density over time, obtained summing over the $xy$ grid cells.}
    %\label{fig:data_generation}
\end{figure}

As a challenging prediction task we consider a setting motivated by the real world problem of modelling bird densities and velocities measured from a set of weather radars \citep{dokter2011bird, nussbaumer2019geostatistical, nussbaumer2021quantifying} -- or more generally the area of radar aeroecology. 
%As a challenging prediction task, 
A non-steady compressible fluid in three dimensions is simulated by propagating fluid parcels through a pre-defined velocity field, i.e. the fluid is simulated using  
%In other words, we simulate a fluid from the Lagrangian perspective with 
the conservation of mass as the underlying PDE (see Eq.~\ref{eq:continuity_equation_masscons}).
%=%=%The paths of the roughly 240'000 parcels are solved for with a basic backward Euler scheme. (added in the appendix)
%=%=%The particle density and velocity are obtained by counting the number of parcels within a voxel, and by averaging the individual velocities. (already explained in the appendix)
To provide the network with training observations, we introduce a set of spatially fixed sensors (comparable to \textit{radars}) which count over time the number of fluid parcels within a radius $r$ and over 21 contiguous altitude layers.
%=%=%Along the $z$-axis, the simulated sensor distinguishes over 21 contiguous altitude layers. (added in the main body)
Another disjoint set of sensors is provided for the validation set while the test performance is evaluated on a grid.
The birds-eye view of the setting is shown in Figure \ref{fig:exp3d_birdeye}, where circles indicate the area covered by the radars.
Figure \ref{fig:exp3d_zaxis} additionally shows the 3D simulated data projected along the $z$-axis and over time. %, which gives a complementary illustration with respect to the birds-eye view.
In the Appendix section \ref{appendix:particle_simulation} we describe the data generation and training setting in detail and provide the corresponding code in the supplementary. % we provide the full code.# for the data generation.

\begin{figure}[ht!]
    \centering
    \includegraphics[width=.95\linewidth]{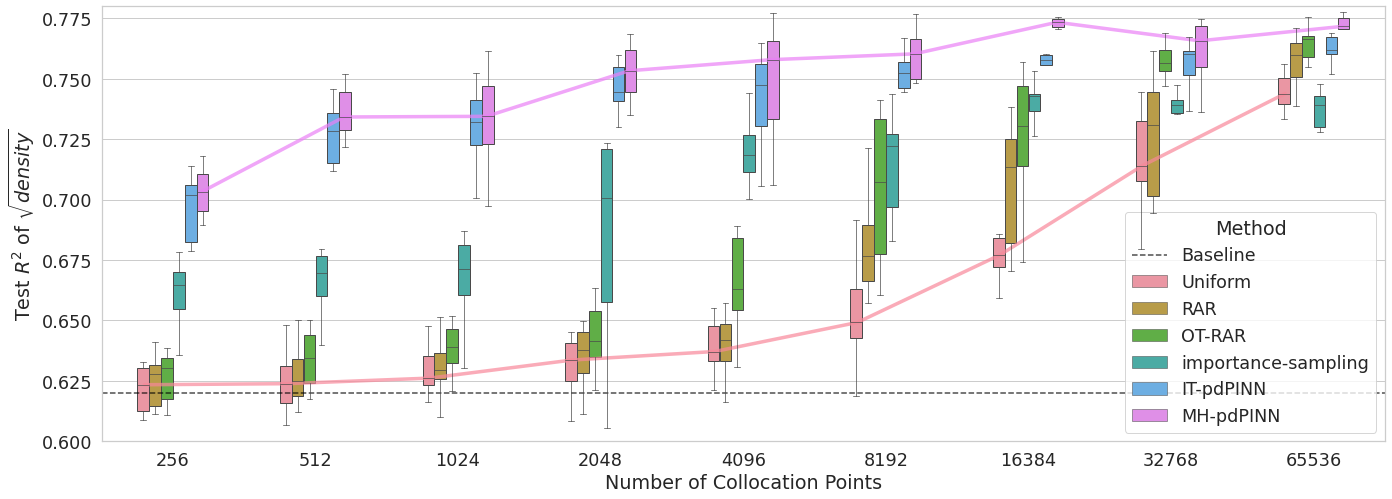}
    \caption{Explained variance of $\sqrt{\rho}$ evaluated on the test set, for different number of collocation points for the 3D mass conservation experiment and for 10 different seeds.}
    \label{fig:exp3d_r2_results}
\end{figure}

For modeling the density and velocity, two sinusoidal representation networks (SIREN) \citep{sitzmann2020implicit} $\rho_{\Theta_1}(t, \bs x)$ and $\bs v_{\Theta_2}(t, \bs x)$ are used, which are then regularized by enforcing the continuity equation for the conservation of mass (see Eq.~\ref{eq:continuity_equation_masscons}).
%=%=%In the 3D setting, data measurements of density and velocity are obtained by $15^2$ sensors on the $xy$-plane, within region $[-3, 3]^2$ at 11 equidistant timesteps. 
To showcase the sample efficiency of pdPINNs, experiments are performed over a wide range of collocation points (256 to 65536).
In each setting the PDE-weights  $w_2$ (see Eq.~\ref{eq:PINN_argmin}) were selected with a grid search based on the highest 1st quartile R$^2$ in a validation set.
%For each method (and each number of collocation points) the used PDE weight was selected based on the highest 1st quartil R$^2$ in a validation set.
The resulting box-plots of the test R$^2$ are provided in Figure \ref{fig:exp3d_r2_results}, where the ``Baseline'' corresponds to training without any PDE loss.
The proposed pdPINN approach clearly outperforms alternative (re-)sampling methods across all numbers of collocation points. %, with importance sampling being the next best competitor.
Already with very few collocation points (512) pdPINNs achieve results that require orders of magnitude more points (32768) for uniform sampling.
Finally, we observe that the performance gap shrinks as the number of collocation points increases, eventually converging to the same limiting value.
Even when getting close to the memory limit of a NVIDIA Titan X GPU, other sampling strategies at best achieve comparable results with pdPINNs.
In the Appendix (Figure \ref{fig:exp3d_proj1d}) we provide an additional qualitative comparison of the mass conservation between OT-RAR and MH-pdPINN 2048 samples.

%We first evaluate the proposed approach qualitatively in terms of density predictions over time, which we illustrate in the Appendix in Figure \ref{fig:exp3d_proj1d}.
%These qualitative results already show that the OT-RAR predictions clearly violate mass conservation while the pdPINN model accurately predicts the density and also faithfully enforcing the continuity equation.
%Qualitative results are reported in the Appendix in Figure \ref{fig:exp3d_proj1d}, displaying the density predictions over time and projected over the $z$-axis.
%This is obtained by summing over the $xy$ grid cells and allows us to visualize that the OT-RAR predictions clearly violate mass conservation.
%Conversely, the pdPINN model accurately predicts the density while also faithfully enforcing the continuity equation.
%In Figure \ref{fig:exp3d_r2_results} we instead evaluate quantitatively the performance of the different methods for several numbers of collocation points.
%For each method, we perform a hyper-parameter search over different PDE-weights $w_2$ (see Eq.~\ref{eq:PINN_argmin}) and select the one with the highest average validation R$^2$.
%The results obtained clearly show that the proposed pdPINN approach outperforms baseline methods across all number of collocation points.
%Already with very few collocation points (512) pdPINNs achieve results comparable with those obtained with orders of magnitude more points (32768).

% Due to the higher dimensionality, the difference between our proposed pdPINN and competing methods is more pronounced.
% Qualitative results are reported in Figure \ref{fig:exp3d_proj1d}, displaying the density predictions over time and projected over the $z$-axis.
%This is obtained by summing over the $xy$ grid cells and allows us to visualize that the OT-RAR predictions clearly violate mass conservation.
%Conversely, the pdPINN model accurately predicts the density while also faithfully enforcing the continuity equation.

%As expected, in this higher-dimensional (3D and time) setting the difference in performances is even more pronounced.
%Already with very few collocation points (512) pdPINNs achieve results comparable with those obtained with orders of magnitude more points (32768).
%We verified that, even when getting close to the memory limit of a NVIDIA Titan X GPU, other sampling strategies at best achieve comparable results with pdPINNs.
%Lastly, results from 2D and 3D experiments suggest the strength of the proposed method would be even more considerable for higher-dimensional settings.
As an additional experiment we simplified the setting by projecting the data onto the \textit{xy}-axis, i.e. the birds-eye view, which is a common setting for geostatistical data (e.g. in \citet{nussbaumer2019geostatistical}).
The results in this 2D setting, which are provided in the Appendix (Figure \ref{fig:exp2d_r2_results}) and described in details in section \ref{appendix:particle_simulation}, are very similar in nature to the 3D setting, although with a smaller performance gap with respect to alternative sampling methods.
This decrease of the gap is to be expected, as the lower dimensional space is much easier to explore with uniform proposals.

\subsection{Heat Equation}
\label{subsec:exp_heat_eq}
\begin{figure}[ht!]
  \centering
    \begin{subfigure}{.22\textwidth}
      \centering
        \includegraphics[width=1.01\linewidth]{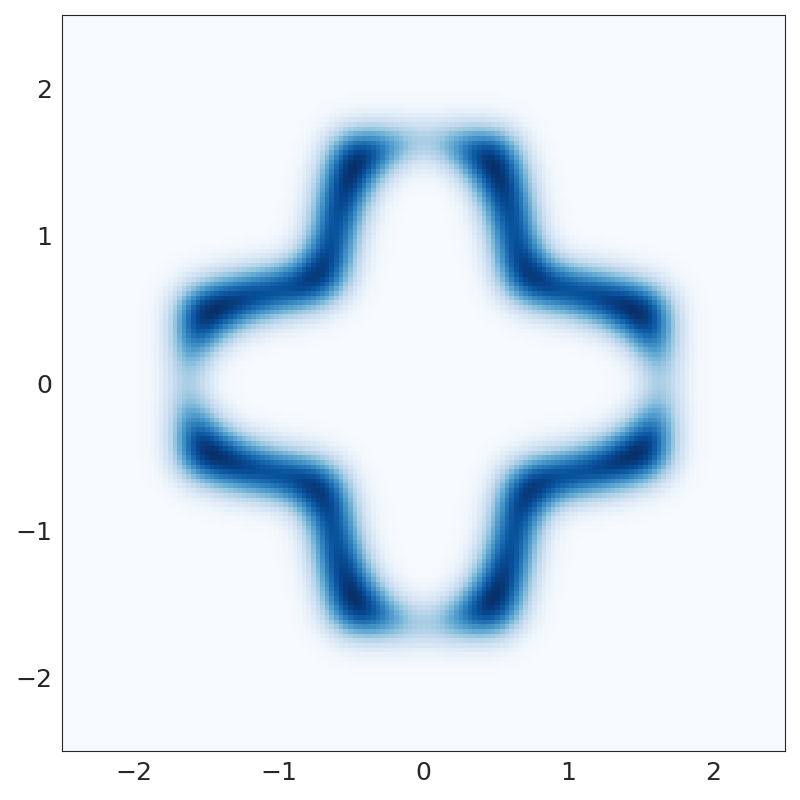}
      \caption{}
      \label{fig:heateq_gt}
    \end{subfigure}%
    \begin{subfigure}{.22\textwidth}
      \centering
        \includegraphics[width=1.01\linewidth]{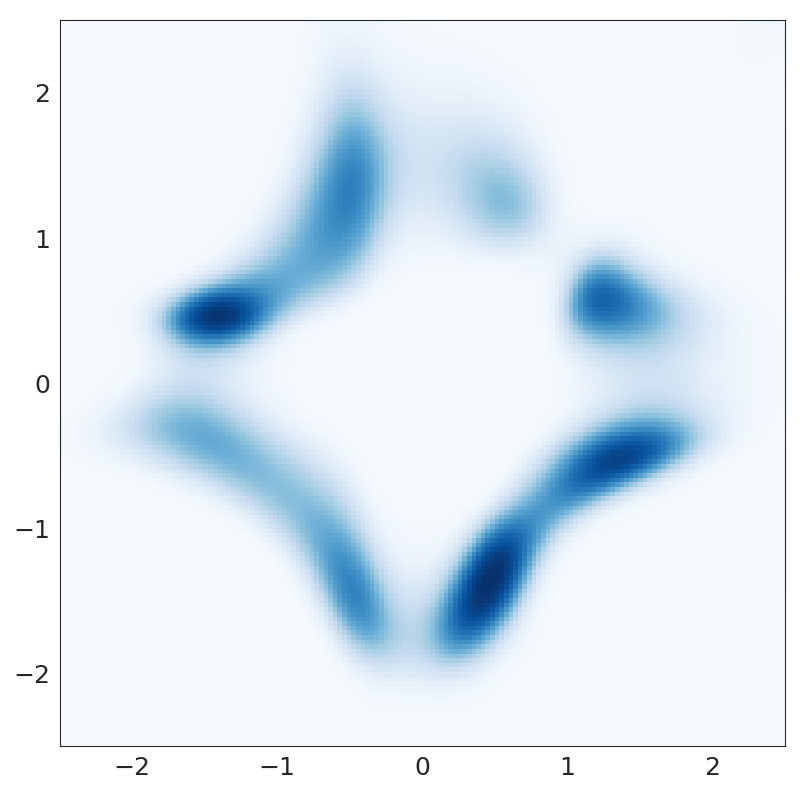}
      \caption{}
      \label{fig:heateq_uniform}
    \end{subfigure}
    \begin{subfigure}{.26\textwidth}
      \centering
        \includegraphics[width=.97\linewidth]{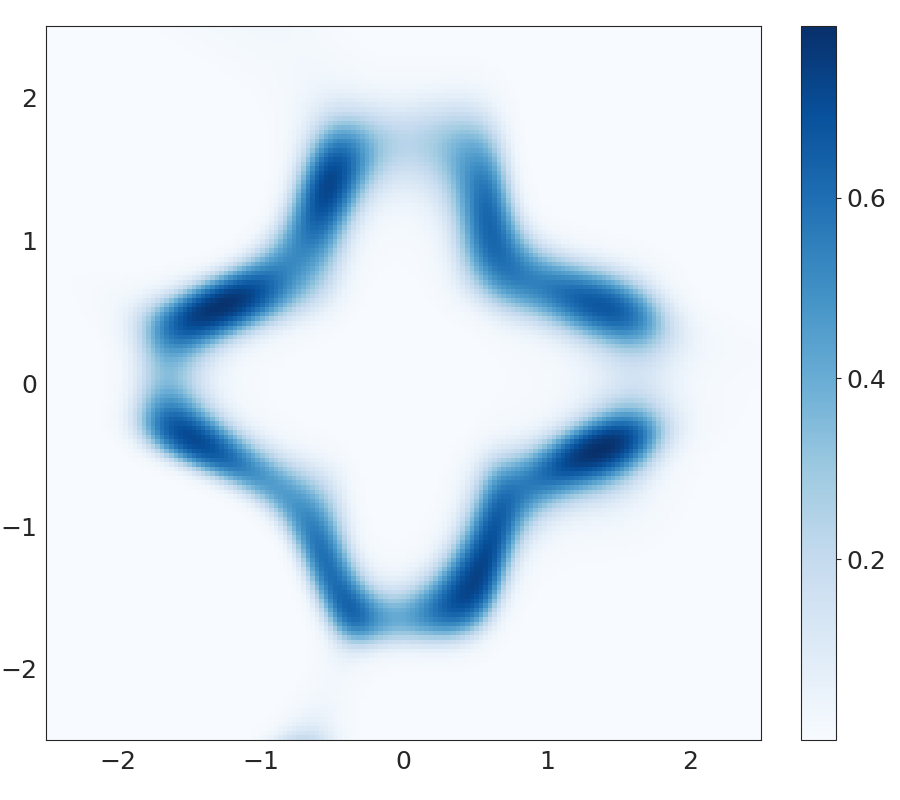}
      \caption{}
      \label{fig:heateq_mcmc}
    \end{subfigure}
  \caption{Temperature predictions of the heat equation experiment (trained with 128 collocation points) at time $t\sim0.044$. 
  (a) Ground truth (b) uniform sampling, and (c) pdPINN.}
\end{figure}

We further consider a 2D diffusion problem, namely the heat equation introduced in section \ref{sec:methodology}, where randomly distributed sensors provide measurements of the temperature. 
%In general, the heat equation is solved in the context of a \textit{Cauchy} problem, where initial and/or boundary conditions can be specified.
%Here 
%We focus on a general setting, where initial conditions are known and are defined by an initial non-zero temperature on the domain.
%Specifically, we focus on the 2-dimensional case and set zero temperature everywhere except for a specified region of the domain, as shown in Figure \hl{CITE}.
We focus on a general setting with the initial conditions being zero temperature everywhere except for a specified region, as shown in Figure \ref{fig:heateq_gt}, and we let the system evolve for $t\in[0,0.2]$.
The networks are only provided sensor measurements of the temperature; for further details see the Appendix section \ref{appendix:heat}.
%This shape allows to break the rotational symmetry of the problem, which would have otherwise reduced the setting to a simpler 1-dimensional problem.
%We compare two methods for selecting the collocation points: 
%A basic PINN implementation with Uniform sampling, and the proposed pdPINN with MCMC sampling from the density.%Markov chain Monte Carlo (MCMC), 
%as described in Section \ref{sec:model_and_implementation}.
%During training we first sample collocation points uniformly, as a warm-up phase, and then re-sample new points every 500 epochs according to the specified sampling strategy.
%=%=%During the warm-up phase of the pdPINN, collocation points are sampled uniformly, and afterwards $90\%$ of the samples are drawn from the particle density distribution, which is proportional to the modeled temperature.
%=%=%Collocation points are re-sampled every 500 epochs.
%=%=%Differently from previous experiments, the employed architecture is a fully-connected two-layer neural network with 32 hidden units and tanh activations. (all three moved to the appendix)

Temperature predictions for PINNs with uniform sampling and pdPINNs are illustrated in Figure \ref{fig:heateq_uniform} and \ref{fig:heateq_mcmc}, respectively, with the ground truth in Figure \ref{fig:heateq_gt}.
We can observe that the uniform sampling strategy does not allow to focus on the relevant parts of the domain, i.e. regions with high temperature, and that it visibly fails to reconstruct the temperature profile.
In contrast, the pdPINN promotes sampling in regions of higher density and predicts the true temperature more reliably.
We also evaluate quantitatively the performance of the two approaches in terms of the R$^2$ test error over the predicted temperature and illustrate the results in the Appendix section \ref{appendix:heat}, where we again observe the same convergence between uniform sampling and pdPINNs for high numbers of collocation points. 
%For the sake of brevity, we provide the results in the Appendix section \ref{appendix:heat}.
%Consistent with previous 2D and 3D compressible fluids experiments, we show that the proposed pdPINNs outperform PINNs with uniform sampling strategy, and that for high numbers of collocation points the performance converges.
%=%=%As in previous settings, we show that with few samples the regularization enforced by the PDE loss is not strong enough, leading to comparable results in both approaches.
%=%=%However, as the number of sample increases, the PDE loss enforced with MCMC samples quickly and steadily outperforms uniform sampling.
%=%=%We also verified that in the limit of high samples the two sampling strategies converge, as expected. (all three integrated in the appendix)
%=%=%Further details regarding the quantitative results are discussed in the Appendix section \ref{appendix:heat}.
%

\subsection{Fokker-Planck Equation}
\label{subsec:exp_fp}
For a demonstration of a forward problem, i.e. a setting without any observed data but only initial conditions, we solve the Fokker-Planck (FP) equations in a setting where an analytical solution is available (cf. \citet{sarkka2019applied}). 
%We showcase the different training behaviours of Uniform sampling, RAR, and
%multiple MCMC samplers in a setting where uniform sampling would already be sufficient to solve
%the task
%Finally, we consider the Fokker-Planck (FP) equations for the solution of a forward problem, given initial conditions.
The FP equations describe the evolution of the probability density of the movement of Brownian particles under a drift.
%
%\begin{figure}[ht!]
%\centering\includegraphics[width=.4\textwidth]{images/exp_fp/comparison_1d.png}
%\caption{KL divergence between the true target distribution and approximation for the 1D Fokker-Planck equation during training for different (re-)sampling methods.}
%\label{figure:fp_1d_img}
%\end{figure}
%
\begin{figure}[ht!]
    \centering
    \begin{subfigure}{.49\textwidth}
      \flushleft
        \includegraphics[width=\linewidth]{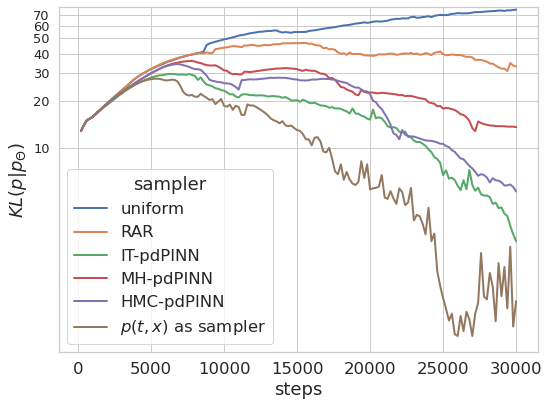}
        \caption{}
        \label{figure:fp_1d_img_maintext}
    \end{subfigure}%
    \hfill
    \begin{subfigure}{.49\textwidth}
        \flushright
        \includegraphics[width=\linewidth]{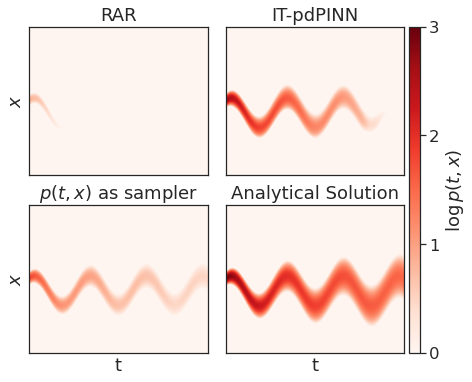}
       \caption{}
       \label{fig:fp_1d_comparison}
    \end{subfigure}%
    \caption{Fokker-Planck equation in 1D. (a) KL divergence between the true target distribution and approximation during training, (b) predicted $\log p_\Theta(t,x)$ after training, cropped to $x\in [-0.5, 0.5]$. }
    %\label{fig:data_generation}
\end{figure}
More specifically, assume we are given particles at time $t_0$, which are distributed according to $p(t_0, x)$.
Let the movements of these particles be described by the following stochastic differential equation, where $W_t$ denotes the standard Wiener process:
\begin{equation}
    dX_t = \mu(t, X_t) \,dt + \sigma(t, X_t) \,dW_t 
\end{equation}
with known drift $\mu(X_t, t)$ and diffusion coefficient $D(X_t, t) = \sigma^2(X_t, t)/2$. 
The FP equation for the probability density $p(t, x)$ of the random variable $X_t$ is then given by
\begin{equation}
\label{eq:fp_1d}
    \frac{\partial}{\partial t} p(t, x) = -\frac{\partial}{\partial x}\left[\mu(t, x) p(t, x)\right] + \frac{\partial^2}{\partial x^2}\left[D(t, x) p(t, x)\right] .
\end{equation}
We train a network to predict the (probability) density $p_\Theta(t, x)$ given a known sinusoidal drift and constant diffusion, which are discussed in detail in the Appendix.
Data is only provided for the initial condition, and the PDE loss is based on Eq.~\ref{eq:fp_1d} within the space $\Omega = [-.1.5, 1.5]$ and time $t\in [-1, 1]$. 
As the analytical solution is available in form of a probability density, we can estimate the KL divergence $KL(p||p_\Theta)$ to evaluate the performance. 
Furthermore, we can sample collocation points from the true particle distribution $p(t,x)$  (referred to as ``$p(t,x)$ as sampler''), offering a ``best case scenario'' of pdPINNs.
A total of 5000 collocation points were used, and weights were manually tuned based on the error on a validation set.
Figure \ref{figure:fp_1d_img_maintext} shows the evolution of KL divergence during training, highlighting that pdPINN based methods require fewer steps to achieve a low divergence.
In addition, sampling from the true particle distribution leads to the fastest improvement and the lowest divergence after 30000 training steps.
A qualitative comparison of the results is given in Figure \ref{fig:fp_1d_comparison}, showing that RAR and uniform sampling fail to propagate the sine wave forward.
The ground truth of the problem and  wall-times for different methods are given in the Appendix section \ref{appendix:fp}.

\section{Conclusion}
\label{sec:conclusion}
In this work, we introduced a general extension to PINNs applicable to a great variety of problem settings 
involving physics-based regularization of neural networks. %architectures.
In order to overcome the limitations of classical mesh-based Eulerian PINNs, we introduce a novel PDE loss that is defined with respect to the particle density in rather general types of PDEs.
By employing MCMC methods to sample collocation points from the density approximated by the network, we derive an efficient and easy-to-implement improvement for providing a more 
appropriate regularization objective in PINNs.  
In particular, our new pdPINNs are completely mesh-free, thereby overcoming severe efficiency problems of classical PINNs in high-dimensional and sparse settings. 
Further, the absence of a mesh allows us to elegantly handle settings with uncertain or unknown domain boundaries.

%These advantages of the pdPINN model increase the relevance of PINNs in higher dimensional real-world problems. 
%Limitations of the proposed approach are settings where the physical variable of interest does not align with the normalized particle density we sample from.
%An example would be modelling incompressible fluids where the focus is on the dynamics of the velocity field rather than on the density, which is constant by assumption.
As we have demonstrated, our method is applicable to a wide spectrum of PDEs, ranging from hydrodynamic flow problems to electro- and thermo-dynamic problems, as well as more general applications of the Fokker-Planck equations.
%Extensions of pdPINNs are possble in 
%Extensions of pdPINNs are possible in more refined sampling procedures.
\vfill

%\begin{ack}
%Use unnumbered first level headings for the acknowledgments. All acknowledgments
%go at the end of the paper before the list of references. Moreover, you are required to declare
%funding (financial activities supporting the submitted work) and competing interests (related financial %activities outside the submitted work).
%More information about this disclosure can be found at: \url{https://neurips.cc/Conferences/2022/PaperInformation/FundingDisclosure}.
%%
%
%Do {\bf not} include this section in the anonymized submission, only in the final paper. You can use the \texttt{ack} environment provided in the style file to autmoatically hide this section in the anonymized submission.
%\end{ack}
\bibliography{main}

\begin{thebibliography}{50}
\providecommand{\natexlab}[1]{#1}
\providecommand{\url}[1]{\texttt{#1}}
\expandafter\ifx\csname urlstyle\endcsname\relax
  \providecommand{\doi}[1]{doi: #1}\else
  \providecommand{\doi}{doi: \begingroup \urlstyle{rm}\Url}\fi

\bibitem[Abadi et~al.(2016)Abadi, Barham, Chen, Chen, Davis, Dean, Devin,
  Ghemawat, Irving, Isard, et~al.]{abadi2016tensorflow}
Abadi, M., Barham, P., Chen, J., Chen, Z., Davis, A., Dean, J., Devin, M.,
  Ghemawat, S., Irving, G., Isard, M., et~al.
\newblock $\{$TensorFlow$\}$: a system for $\{$Large-Scale$\}$ machine
  learning.
\newblock In \emph{12th USENIX symposium on operating systems design and
  implementation (OSDI 16)}, pp.\  265--283, 2016.

\bibitem[Bertin et~al.(2006)Bertin, Droz, and
  Gr{\'e}goire]{bertin2006boltzmann}
Bertin, E., Droz, M., and Gr{\'e}goire, G.
\newblock Boltzmann and hydrodynamic description for self-propelled particles.
\newblock \emph{Physical Review E}, 74\penalty0 (2):\penalty0 022101, 2006.

\bibitem[Betancourt(2017)]{betancourt2017conceptual}
Betancourt, M.
\newblock A conceptual introduction to hamiltonian monte carlo.
\newblock \emph{arXiv preprint arXiv:1701.02434}, 2017.

\bibitem[Born \& Green(1946)Born and Green]{born1946general}
Born, M. and Green, H.~S.
\newblock A general kinetic theory of liquids i. the molecular distribution
  functions.
\newblock \emph{Proceedings of the Royal Society of London. Series A.
  Mathematical and Physical Sciences}, 188\penalty0 (1012):\penalty0 10--18,
  1946.

\bibitem[Chen et~al.(2019)Chen, Du, Li, and Lyu]{chen2019quasi}
Chen, J., Du, R., Li, P., and Lyu, L.
\newblock Quasi-monte carlo sampling for machine-learning partial differential
  equations.
\newblock \emph{arXiv preprint arXiv:1911.01612}, 2019.

\bibitem[Dissanayake \& Phan-Thien(1994)Dissanayake and
  Phan-Thien]{dissanayake1994neural}
Dissanayake, M. and Phan-Thien, N.
\newblock Neural-network-based approximations for solving partial differential
  equations.
\newblock \emph{communications in Numerical Methods in Engineering},
  10\penalty0 (3):\penalty0 195--201, 1994.

\bibitem[Dokter et~al.(2011)Dokter, Liechti, Stark, Delobbe, Tabary, and
  Holleman]{dokter2011bird}
Dokter, A.~M., Liechti, F., Stark, H., Delobbe, L., Tabary, P., and Holleman,
  I.
\newblock Bird migration flight altitudes studied by a network of operational
  weather radars.
\newblock \emph{Journal of the Royal Society Interface}, 8\penalty0
  (54):\penalty0 30--43, 2011.

\bibitem[Duane et~al.(1987)Duane, Kennedy, Pendleton, and
  Roweth]{duane1987hybrid}
Duane, S., Kennedy, A.~D., Pendleton, B.~J., and Roweth, D.
\newblock Hybrid monte carlo.
\newblock \emph{Physics letters B}, 195\penalty0 (2):\penalty0 216--222, 1987.

\bibitem[Earl \& Deem(2005)Earl and Deem]{earl2005parallel}
Earl, D.~J. and Deem, M.~W.
\newblock Parallel tempering: Theory, applications, and new perspectives.
\newblock \emph{Physical Chemistry Chemical Physics}, 7\penalty0 (23):\penalty0
  3910--3916, 2005.

\bibitem[Flamary et~al.(2021)Flamary, Courty, Gramfort, Alaya, Boisbunon,
  Chambon, Chapel, Corenflos, Fatras, Fournier, Gautheron, Gayraud, Janati,
  Rakotomamonjy, Redko, Rolet, Schutz, Seguy, Sutherland, Tavenard, Tong, and
  Vayer]{flamary2021pot}
Flamary, R., Courty, N., Gramfort, A., Alaya, M.~Z., Boisbunon, A., Chambon,
  S., Chapel, L., Corenflos, A., Fatras, K., Fournier, N., Gautheron, L.,
  Gayraud, N.~T., Janati, H., Rakotomamonjy, A., Redko, I., Rolet, A., Schutz,
  A., Seguy, V., Sutherland, D.~J., Tavenard, R., Tong, A., and Vayer, T.
\newblock Pot: Python optimal transport.
\newblock \emph{Journal of Machine Learning Research}, 22\penalty0
  (78):\penalty0 1--8, 2021.

\bibitem[Freitag(2020)]{freitag2020numerical}
Freitag, M.~A.
\newblock Numerical linear algebra in data assimilation.
\newblock \emph{GAMM-Mitteilungen}, 43\penalty0 (3):\penalty0 e202000014, 2020.

\bibitem[Gingold \& Monaghan(1977)Gingold and Monaghan]{gingold1977smoothed}
Gingold, R.~A. and Monaghan, J.~J.
\newblock Smoothed particle hydrodynamics: theory and application to
  non-spherical stars.
\newblock \emph{Monthly notices of the royal astronomical society},
  181\penalty0 (3):\penalty0 375--389, 1977.

\bibitem[Green(1956)]{green1956boltzmann}
Green, M.~S.
\newblock Boltzmann equation from the statistical mechanical point of view.
\newblock \emph{The Journal of Chemical Physics}, 25\penalty0 (5):\penalty0
  836--855, 1956.

\bibitem[Hoffman et~al.(2014)Hoffman, Gelman, et~al.]{hoffman2014no}
Hoffman, M.~D., Gelman, A., et~al.
\newblock The no-u-turn sampler: adaptively setting path lengths in hamiltonian
  monte carlo.
\newblock \emph{J. Mach. Learn. Res.}, 15\penalty0 (1):\penalty0 1593--1623,
  2014.

\bibitem[Hoover \& Hoover(2003)Hoover and Hoover]{hoover2003links}
Hoover, W.~G. and Hoover, C.
\newblock Links between microscopic and macroscopic fluid mechanics.
\newblock \emph{Molecular Physics}, 101\penalty0 (11):\penalty0 1559--1573,
  2003.

\bibitem[Jagtap et~al.(2020)Jagtap, Kharazmi, and
  Karniadakis]{jagtap2020conservative}
Jagtap, A.~D., Kharazmi, E., and Karniadakis, G.~E.
\newblock Conservative physics-informed neural networks on discrete domains for
  conservation laws: Applications to forward and inverse problems.
\newblock \emph{Computer Methods in Applied Mechanics and Engineering},
  365:\penalty0 113028, 2020.

\bibitem[Kingma \& Ba(2014)Kingma and Ba]{kingma2016adam}
Kingma, D.~P. and Ba, J.
\newblock Adam: A method for stochastic optimization.
\newblock \emph{arXiv preprint arXiv:1412.6980}, 2014.

\bibitem[Knott \& Smith(1984)Knott and Smith]{knott1984optimal}
Knott, M. and Smith, C.~S.
\newblock On the optimal mapping of distributions.
\newblock \emph{Journal of Optimization Theory and Applications}, 43\penalty0
  (1):\penalty0 39--49, 1984.

\bibitem[Lagaris et~al.(1998)Lagaris, Likas, and
  Fotiadis]{lagaris1998artificial}
Lagaris, I.~E., Likas, A., and Fotiadis, D.~I.
\newblock Artificial neural networks for solving ordinary and partial
  differential equations.
\newblock \emph{IEEE transactions on neural networks}, 9\penalty0 (5):\penalty0
  987--1000, 1998.

\bibitem[Lagaris et~al.(2000)Lagaris, Likas, and
  Papageorgiou]{lagaris2000neural}
Lagaris, I.~E., Likas, A.~C., and Papageorgiou, D.~G.
\newblock Neural-network methods for boundary value problems with irregular
  boundaries.
\newblock \emph{IEEE Transactions on Neural Networks}, 11\penalty0
  (5):\penalty0 1041--1049, 2000.

\bibitem[Lao et~al.(2020)Lao, Suter, Langmore, Chimisov, Saxena, Sountsov,
  Moore, Saurous, Hoffman, and Dillon]{lao2020tfp}
Lao, J., Suter, C., Langmore, I., Chimisov, C., Saxena, A., Sountsov, P.,
  Moore, D., Saurous, R.~A., Hoffman, M.~D., and Dillon, J.~V.
\newblock tfp. mcmc: Modern markov chain monte carlo tools built for modern
  hardware.
\newblock \emph{arXiv preprint arXiv:2002.01184}, 2020.

\bibitem[Lind et~al.(2020)Lind, Rogers, and Stansby]{lind2020review}
Lind, S.~J., Rogers, B.~D., and Stansby, P.~K.
\newblock Review of smoothed particle hydrodynamics: towards converged
  lagrangian flow modelling.
\newblock \emph{Proceedings of the Royal Society A}, 476\penalty0
  (2241):\penalty0 20190801, 2020.

\bibitem[Lu et~al.(2021)Lu, Meng, Mao, and Karniadakis]{lu2021deepxde}
Lu, L., Meng, X., Mao, Z., and Karniadakis, G.~E.
\newblock Deepxde: A deep learning library for solving differential equations.
\newblock \emph{SIAM Review}, 63\penalty0 (1):\penalty0 208--228, 2021.

\bibitem[Marchetti et~al.(2013)Marchetti, Joanny, Ramaswamy, Liverpool, Prost,
  Rao, and Simha]{marchetti2013hydrodynamics}
Marchetti, M.~C., Joanny, J.-F., Ramaswamy, S., Liverpool, T.~B., Prost, J.,
  Rao, M., and Simha, R.~A.
\newblock Hydrodynamics of soft active matter.
\newblock \emph{Reviews of modern physics}, 85\penalty0 (3):\penalty0 1143,
  2013.

\bibitem[Meurer et~al.(2017)Meurer, Smith, Paprocki, \v{C}ert\'{i}k, Kirpichev,
  Rocklin, Kumar, Ivanov, Moore, Singh, Rathnayake, Vig, Granger, Muller,
  Bonazzi, Gupta, Vats, Johansson, Pedregosa, Curry, Terrel, Rou\v{c}ka, Saboo,
  Fernando, Kulal, Cimrman, and Scopatz]{sympy}
Meurer, A., Smith, C.~P., Paprocki, M., \v{C}ert\'{i}k, O., Kirpichev, S.~B.,
  Rocklin, M., Kumar, A., Ivanov, S., Moore, J.~K., Singh, S., Rathnayake, T.,
  Vig, S., Granger, B.~E., Muller, R.~P., Bonazzi, F., Gupta, H., Vats, S.,
  Johansson, F., Pedregosa, F., Curry, M.~J., Terrel, A.~R., Rou\v{c}ka, v.,
  Saboo, A., Fernando, I., Kulal, S., Cimrman, R., and Scopatz, A.
\newblock Sympy: symbolic computing in python.
\newblock \emph{PeerJ Computer Science}, 3:\penalty0 e103, January 2017.
\newblock ISSN 2376-5992.
\newblock \doi{10.7717/peerj-cs.103}.

\bibitem[Nabian et~al.(2021)Nabian, Gladstone, and
  Meidani]{nabian2021efficient}
Nabian, M.~A., Gladstone, R.~J., and Meidani, H.
\newblock Efficient training of physics-informed neural networks via importance
  sampling.
\newblock \emph{Computer-Aided Civil and Infrastructure Engineering},
  36\penalty0 (8):\penalty0 962--977, 2021.

\bibitem[Nussbaumer et~al.(2019)Nussbaumer, Benoit, Mariethoz, Liechti, Bauer,
  and Schmid]{nussbaumer2019geostatistical}
Nussbaumer, R., Benoit, L., Mariethoz, G., Liechti, F., Bauer, S., and Schmid,
  B.
\newblock A geostatistical approach to estimate high resolution nocturnal bird
  migration densities from a weather radar network.
\newblock \emph{Remote Sensing}, 11\penalty0 (19):\penalty0 2233, 2019.

\bibitem[Nussbaumer et~al.(2021)Nussbaumer, Bauer, Benoit, Mariethoz, Liechti,
  and Schmid]{nussbaumer2021quantifying}
Nussbaumer, R., Bauer, S., Benoit, L., Mariethoz, G., Liechti, F., and Schmid,
  B.
\newblock Quantifying year-round nocturnal bird migration with a fluid dynamics
  model.
\newblock \emph{Journal of the Royal Society Interface}, 18\penalty0
  (179):\penalty0 20210194, 2021.

\bibitem[Oosthuizen \& Carscallen(2013)Oosthuizen and
  Carscallen]{oosthuizen2013introduction}
Oosthuizen, P.~H. and Carscallen, W.~E.
\newblock \emph{Introduction to compressible fluid flow}.
\newblock CRC press, 2013.

\bibitem[Parisi et~al.(2003)Parisi, Mariani, and Laborde]{parisi2003solving}
Parisi, D.~R., Mariani, M.~C., and Laborde, M.~A.
\newblock Solving differential equations with unsupervised neural networks.
\newblock \emph{Chemical Engineering and Processing: Process Intensification},
  42\penalty0 (8-9):\penalty0 715--721, 2003.

\bibitem[Parno \& Marzouk(2018)Parno and Marzouk]{parno2018transport}
Parno, M.~D. and Marzouk, Y.~M.
\newblock Transport map accelerated markov chain monte carlo.
\newblock \emph{SIAM/ASA Journal on Uncertainty Quantification}, 6\penalty0
  (2):\penalty0 645--682, 2018.

\bibitem[Paszke et~al.(2019)Paszke, Gross, Massa, Lerer, Bradbury, Chanan,
  Killeen, Lin, Gimelshein, Antiga, Desmaison, Kopf, Yang, DeVito, Raison,
  Tejani, Chilamkurthy, Steiner, Fang, Bai, and Chintala]{pytorch}
Paszke, A., Gross, S., Massa, F., Lerer, A., Bradbury, J., Chanan, G., Killeen,
  T., Lin, Z., Gimelshein, N., Antiga, L., Desmaison, A., Kopf, A., Yang, E.,
  DeVito, Z., Raison, M., Tejani, A., Chilamkurthy, S., Steiner, B., Fang, L.,
  Bai, J., and Chintala, S.
\newblock Pytorch: An imperative style, high-performance deep learning library.
\newblock In Wallach, H., Larochelle, H., Beygelzimer, A., d\textquotesingle
  Alch\'{e}-Buc, F., Fox, E., and Garnett, R. (eds.), \emph{Advances in Neural
  Information Processing Systems 32}, pp.\  8024--8035. Curran Associates,
  Inc., 2019.

\bibitem[Raissi et~al.(2019)Raissi, Perdikaris, and
  Karniadakis]{raissi2019physics}
Raissi, M., Perdikaris, P., and Karniadakis, G.~E.
\newblock Physics-informed neural networks: A deep learning framework for
  solving forward and inverse problems involving nonlinear partial differential
  equations.
\newblock \emph{Journal of Computational Physics}, 378:\penalty0 686--707,
  2019.

\bibitem[Recktenwald(2004)]{Recktenwald2004FiniteDifferenceAT}
Recktenwald, G.~W.
\newblock Finite-difference approximations to the heat equation.
\newblock \emph{Mechanical Engineering}, 10\penalty0 (01), 2004.

\bibitem[Rezende \& Mohamed(2015)Rezende and Mohamed]{rezende2015variational}
Rezende, D. and Mohamed, S.
\newblock Variational inference with normalizing flows.
\newblock In \emph{International conference on machine learning}, pp.\
  1530--1538. PMLR, 2015.

\bibitem[Rudd(2013)]{rudd2013solving}
Rudd, K.
\newblock \emph{Solving partial differential equations using artificial neural
  networks}.
\newblock PhD thesis, Duke University, 2013.

\bibitem[S{\"a}rkk{\"a} \& Solin(2019)S{\"a}rkk{\"a} and
  Solin]{sarkka2019applied}
S{\"a}rkk{\"a}, S. and Solin, A.
\newblock \emph{Applied stochastic differential equations}, volume~10.
\newblock Cambridge University Press, 2019.

\bibitem[Sirignano \& Spiliopoulos(2018)Sirignano and
  Spiliopoulos]{sirignano2018dgm}
Sirignano, J. and Spiliopoulos, K.
\newblock Dgm: A deep learning algorithm for solving partial differential
  equations.
\newblock \emph{Journal of computational physics}, 375:\penalty0 1339--1364,
  2018.

\bibitem[Sitzmann et~al.(2020)Sitzmann, Martel, Bergman, Lindell, and
  Wetzstein]{sitzmann2020implicit}
Sitzmann, V., Martel, J., Bergman, A., Lindell, D., and Wetzstein, G.
\newblock Implicit neural representations with periodic activation functions.
\newblock \emph{Advances in Neural Information Processing Systems},
  33:\penalty0 7462--7473, 2020.

\bibitem[Steele(1987)]{steele1987non}
Steele, J.~M.
\newblock Non-uniform random variate generation (luc devroye), 1987.

\bibitem[Szabó et~al.(2006)Szabó, Sz{\"o}ll{\"o}si, G{\"o}nci, Jur{\'a}nyi,
  Selmeczi, and Vicsek]{szabo2006phase}
Szabó, B., Sz{\"o}ll{\"o}si, G., G{\"o}nci, B., Jur{\'a}nyi, Z., Selmeczi, D.,
  and Vicsek, T.
\newblock Phase transition in the collective migration of tissue cells:
  Experiment and model.
\newblock \emph{Physical Review E}, 74\penalty0 (6):\penalty0 061908, 2006.

\bibitem[Tadiparthi \& Bhattacharya(2021)Tadiparthi and
  Bhattacharya]{tadiparthi2021optimal}
Tadiparthi, V. and Bhattacharya, R.
\newblock Optimal transport based refinement of physics-informed neural
  networks.
\newblock \emph{arXiv preprint arXiv:2105.12307}, 2021.

\bibitem[Tancik et~al.(2020)Tancik, Srinivasan, Mildenhall, Fridovich-Keil,
  Raghavan, Singhal, Ramamoorthi, Barron, and Ng]{tancik2020fourier}
Tancik, M., Srinivasan, P., Mildenhall, B., Fridovich-Keil, S., Raghavan, N.,
  Singhal, U., Ramamoorthi, R., Barron, J., and Ng, R.
\newblock Fourier features let networks learn high frequency functions in low
  dimensional domains.
\newblock \emph{Advances in Neural Information Processing Systems},
  33:\penalty0 7537--7547, 2020.

\bibitem[Toner \& Tu(1995)Toner and Tu]{toner1995long}
Toner, J. and Tu, Y.
\newblock Long-range order in a two-dimensional dynamical {XY} model: how birds
  fly together.
\newblock \emph{Physical review letters}, 75\penalty0 (23):\penalty0 4326,
  1995.

\bibitem[Toner \& Tu(1998)Toner and Tu]{toner1998flocks}
Toner, J. and Tu, Y.
\newblock Flocks, herds, and schools: A quantitative theory of flocking.
\newblock \emph{Physical review E}, 58\penalty0 (4):\penalty0 4828, 1998.

\bibitem[Tu et~al.(1998)Tu, Toner, and Ulm]{tu1998sound}
Tu, Y., Toner, J., and Ulm, M.
\newblock Sound waves and the absence of {G}alilean invariance in flocks.
\newblock \emph{Physical review letters}, 80\penalty0 (21):\penalty0 4819,
  1998.

\bibitem[van Milligen et~al.(1995)van Milligen, Tribaldos, and
  Jim{\'e}nez]{van1995neural}
van Milligen, B.~P., Tribaldos, V., and Jim{\'e}nez, J.
\newblock Neural network differential equation and plasma equilibrium solver.
\newblock \emph{Physical review letters}, 75\penalty0 (20):\penalty0 3594,
  1995.

\bibitem[Wang et~al.(2021{\natexlab{a}})Wang, Teng, and
  Perdikaris]{wang2021understanding}
Wang, S., Teng, Y., and Perdikaris, P.
\newblock Understanding and mitigating gradient flow pathologies in
  physics-informed neural networks.
\newblock \emph{SIAM Journal on Scientific Computing}, 43\penalty0
  (5):\penalty0 A3055--A3081, 2021{\natexlab{a}}.

\bibitem[Wang et~al.(2021{\natexlab{b}})Wang, Wang, and
  Perdikaris]{wang2021eigenvector}
Wang, S., Wang, H., and Perdikaris, P.
\newblock On the eigenvector bias of fourier feature networks: From regression
  to solving multi-scale pdes with physics-informed neural networks.
\newblock \emph{Computer Methods in Applied Mechanics and Engineering},
  384:\penalty0 113938, 2021{\natexlab{b}}.

\bibitem[Wessels et~al.(2020)Wessels, Wei{\ss}enfels, and
  Wriggers]{wessels2020neural}
Wessels, H., Wei{\ss}enfels, C., and Wriggers, P.
\newblock The neural particle method--an updated lagrangian physics informed
  neural network for computational fluid dynamics.
\newblock \emph{Computer Methods in Applied Mechanics and Engineering},
  368:\penalty0 113127, 2020.

\end{thebibliography}

%%%%%%%%%%%%%%%%%%%%%%%%%%%%%%%%%%%%%%%%%%%%%%%%%%%%%%%%%%%%
% Checklist
%\newpage
%\section*{Checklist}
%\input{subfiles/08_checklist}
%%%%%%%%%%%%%%%%%%%%%%%%%%%%%%%%%%%%%%%%%%%%%%%%%%%%%%%%%%%%

\newpage
\appendix
\section{Appendix}
\renewcommand\thefigure{A.\arabic{figure}}
\renewcommand\thetable{A.\arabic{table}}

\subsection{Background Sampling for pdPINNs}
\label{appendix:background}
At initialization, the network prediction $\rho_\Theta$ is random and thus does not carry any useful information, i.e. sampling from this density would be meaningless.
Therefore, we start training the pdPINNs with a warm-up phase in which samples are obtained from a pre-specified background distribution:
\begin{equation}
    \bs x \sim p_{bg}(t, \bs x)=p(t)p_{bg}(\bs x | t)
\end{equation}
with $p(t) = \mathcal{U}(0, T)$.
To avoid introducing a mesh, we could rely on the previously estimated Gaussian distribution introduced in Section \ref{sec:model_and_implementation}, i.e. $p_{bg}(\bs x|t) = p_{\text{gauss}}(\bs x)$.
As a second alternative, approach we consider random linear combinations of the convex hull of $\{x^{(i)}\}_{i=1}^N$ spanned by $c$ data points summarized as rows of matrix $Z\in \mathbb{R}^{c\times d}$. This leads to $\bs x = {\bs m}Z$ with weight $\bs m\in \mathbb{R}^c$ which can be drawn from a Dirichlet distribution, i.e. $\bs m \sim Dir(\bs \alpha=1)$.
%
% \begin{align}
%     \bs m &\sim Dir(\bs \alpha=1) &
%     \bs x &= mZ
% \end{align}
%
Of course, a uniform sampling mechanism on a defined region is also suitable
and the definitive choice depends on the data and PDE at hand.
However, we found that all of these methods work well in practice.

We initially draw all samples from the background distribution, and then slowly increase the proportion of samples obtained from the particle density, as we found that leaving some background samples slightly helps in the training.

\subsection{Implementation of RAR and OT-RAR}
\label{appendix:implementation}
For our comparison, we considered the adaptive refinement methods \textit{RAR} and \textit{OT-RAR}, proposed by \citet{lu2021deepxde} and \citet[preprint]{tadiparthi2021optimal}.
Both methods rely on consecutive refinements of a fixed grid in the initial proposal. 
The number of collocation points is steadily increased and collocation points once added will not be removed.
To allow for a fairer comparison, we adapt both methods to use a limited budget of points, and in addition we regularly resample them. 
This leads to a slightly modified version of the methods which is similar in spirit.
For learning the linear mapping proposed by \citet{tadiparthi2021optimal}, we rely on the PyOT \citep{flamary2021pot} implementation of \citet{knott1984optimal}.
The pseudo-code for sampling a set of collocation points is given in Algorithm \ref{alg:rar} and Algorithm \ref{alg:ot-rar}.
The required input $f_\Theta$ refers to the PDE approximated by the network, as discussed in Section \ref{sec:introduction}.
For more specific details on the methods we refer to the original papers.

%It should be noted, that no code was provided by \citet{tadiparthi2021optimal}, and the selection of the target empirical distribution for learning the linear map is not described in detail.
%As such, our implementation of OT-RAR is to the best of our knowledge.

%\textcolor{magenta}{Reiterate what $f_\Theta$ means and maybe explain $k$//4.}

\begin{algorithm}
\caption{Adapted RAR}\label{alg:rar}
\begin{algorithmic}
\Require $f_\Theta$, uniform distribution $\mathcal{U}_{\mathcal{B}}$,\\ \hspace{7mm}  number of col. points $k$,\  previous col. points $X_{\text{prev}}$.
\\
\State $X_{\text{prop}} \hspace{1.2mm} \gets [\bs x_1,\bs x_2, \dots, \bs x_k]^T$ with $\bs x_i \sim \mathcal{U}_{\mathcal{B}}$ \Comment{Sample proposals}
\State $X_\text{comb} \hspace{0.3mm} \gets$ concat($X_{\text{prev}}$, $X_{\text{prop}}$) \Comment{Concatenate old and new points}
\State $X_{\text{new}} \hspace{1.7mm} \gets \text{topk}(X_\text{comb},\ ||f_\Theta(X_\text{comb})||_2^2,\ k)$ \Comment{Keep top $k$ proposed points based on $f_{\Theta}$}
\\
\Ensure $X_{\text{new}}$
\end{algorithmic}
\end{algorithm}

\begin{algorithm}
\caption{Adapted OT-RAR}
\label{alg:ot-rar}
\begin{algorithmic}
\Require $f_\Theta$, uniform distribution $\mathcal{U}_{\mathcal{B}}$, \\ \hspace{7mm} number of col. points $k$, \\ \hspace{8mm}number of points for empirical distribution $j<2k$, \\ \hspace{7mm} previous col. points $X_{\text{prev}}$.
\\
\State $X_{\text{prop}} \hspace{2mm} \gets [\bs x_1,\bs x_2, \dots, \bs x_k]^T$ with $\bs x_i \sim \mathcal{U}_{\mathcal{B}}$ \Comment{Sample proposals}
\State $X_\text{comb} \hspace{1.1mm} \gets$ concat($X_{\text{prev}}$, $X_{\text{prop}}$) \Comment{Concatenate old and new points}
\\
\State $X_{\text{target}} \hspace{1.1mm} \gets \text{topk}(X_\text{comb},\ ||f_\Theta(X_\text{comb})||_2^2,\ j)$ \Comment{$j$ samples for target empirical distribution}
\State $X_{\text{source}} \hspace{0.3mm} \gets [\bs x_1,\bs x_2, \dots, \bs x_{j}]^T$ with $\bs x_i \sim \mathcal{U}_{\mathcal{B}}$ \Comment{$j$ samples for source empirical distribution}
\\
\State $M_\text{OT} \hspace{2.9mm} \gets \text{LinOT}(X_{\text{source}}, X_{\text{target}})$ \Comment{Obtain linear operator that maps to target distribution}
\\
\State $X_{\text{new}} \hspace{2.5mm} \gets [\bs x_1,\bs x_2, \dots, \bs x_k]^T$ with $\bs x_i \sim \mathcal{U}_{\mathcal{B}}$ \Comment{Sample uniformly}
\State $X_{\text{map}} \hspace{2.3mm} \gets M_\text{OT}(X_{\text{new}}) $ \Comment{Map samples to target distribution}
\\
\Ensure $X_{\text{map}}$
\end{algorithmic}
\end{algorithm}

\subsection{Experiments: Conservation of Mass}
\label{appendix:particle_simulation}
In the supplementary material we provide code in Python for the data generation and for the pdPINN model. Below we provide the details for all the experiments we conducted.
Furthermore, we provide short videos showing the predicted density movements for each different approach. 
More details on this can be found in the \texttt{README.html} %\href{README.html}{README.html}  
provided in the supplementary files.
%The model is implemented using PyTorch \citep{pytorch}, with a custom Python implementation of the Metropolis-Hastings MCMC sampler relying on NumPy, SciPy, and Numba \citep{numpy, scipy, numba}.

All experiments were run on a computing cluster using \textit{Nvidia GeForce GTX Titan X GPUs} with 12 GB VRAM.
Settings that required more memory were run on a RTX8000 with 48GB VRAM.
Up to 16 Titan X GPUs could be used in parallel, or 4 RTX8000. %were available, which were used to parallelize multiple runs.
In most settings, training in each experiment took less than 10 minutes.
%The only exceptions were settings with a large amount of collocation points. %, with the bottleneck being the non-optimized bare-bones implementation of the Metropolis-Hastings sampler.

\subsubsection{Additional Experimental Results}
\begin{figure}[ht!]
    \centering
    \includegraphics[width=\linewidth]{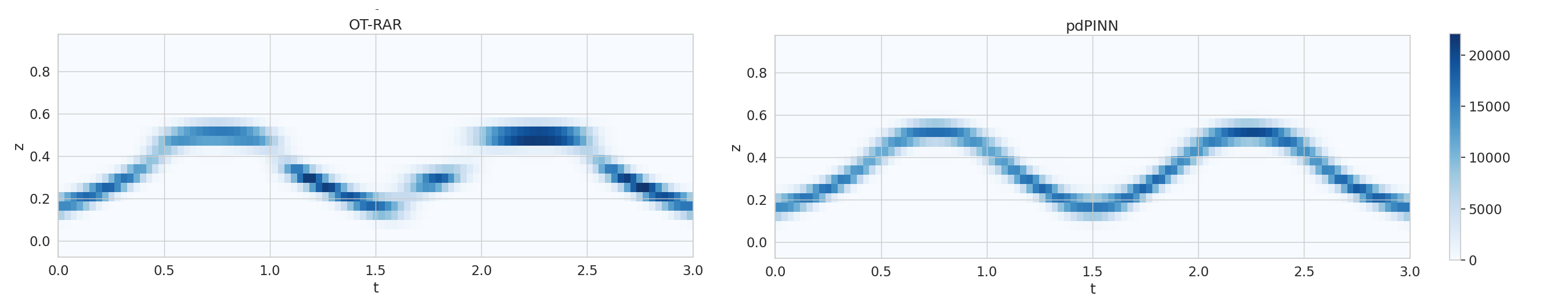}
    \caption{Mass conservation experiment (3D): Predictions (obtained with 2048 collocation points) summed over $xy$ grid cells to obtain $z$-axis projection over time.}
    \label{fig:exp3d_proj1d}
\end{figure}

\begin{figure}[ht!]
    \centering
    \includegraphics[width=.95\linewidth]{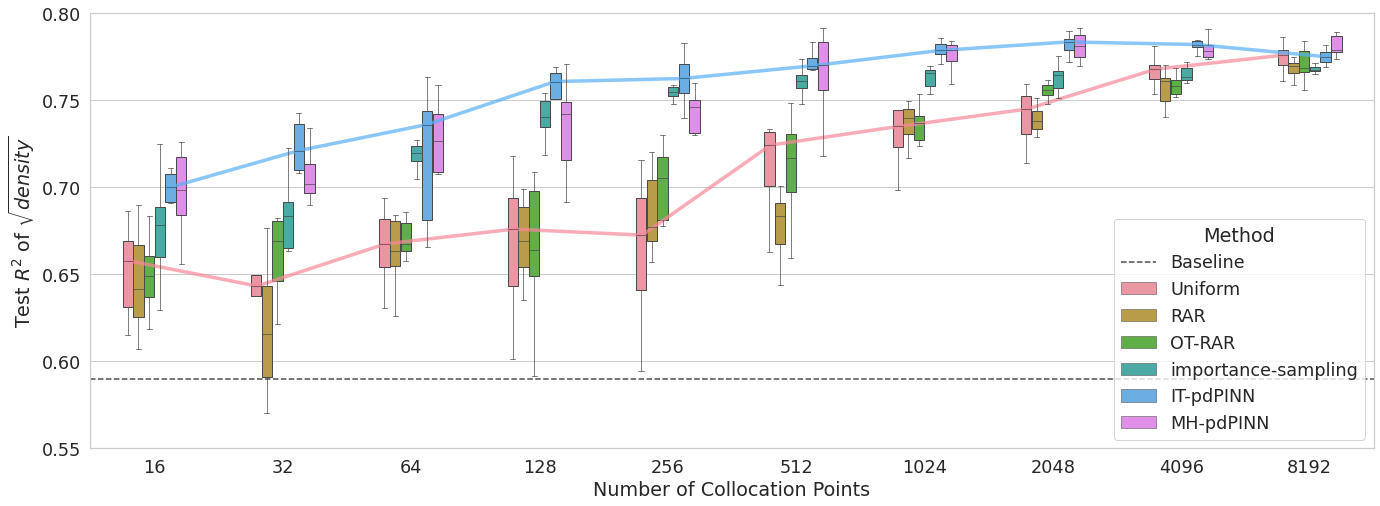}
    \caption{Explained variance of $\sqrt{\rho}$ evaluated on the test set, for different number of collocation points for the 2D mass conservation experiment.
    }
    \label{fig:exp2d_r2_results}
\end{figure}

\paragraph{3D Setting.}
Figure \ref{fig:exp3d_proj1d} showcases the projection of the density in the onto the z axis for a random run of the OT-RAR method and the Metropolis-Hastings based pdPINN when
using 2048 collocation points.
The OT-RAR PINN shows disconnected density predictions that clearly violate mass conservation, whereas the Metropolis Hastings based pdPINN is capable of mostly preserving it.
The boxplot in Figure \ref{fig:exp2d_r2_results} highlights the difference in required number of collocation points of
\begin{figure}[ht!]
      \centering
        \includegraphics[width=.3\linewidth]{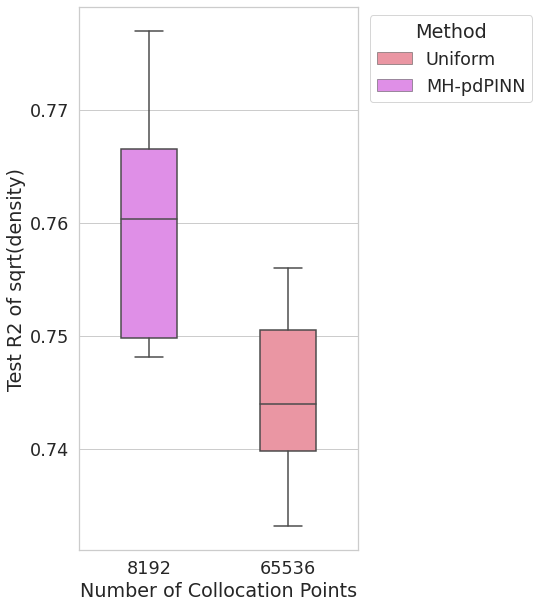}
      \caption{Mass conservation experiment (3D): Boxplot of test R$^2$ of $\sqrt{\rho}$ comparing pdPINN and uniform sampling with a factor 8 difference for the number of collocation points.
      For each method, the used PDE weight was selected based on the highest 1st quartile R$^2$ in a validation set.}
      \label{app:fig:exp3d_8k_vs_65k}
\end{figure}
\paragraph{2D Setting.}
As mentioned in Section \ref{sec:experiments}, we repeated the Conservation of Mass experiment in a slightly altered setting, where the data is projected onto the \textit{xy}-plane, reducing it to a 2D+Time problem.
The general setup is similar to the 3D setting, although a smaller network and different training parameters are used, which are listed in the following sections below.

\subsubsection{Data Generation}
Here we provide a more detailed description for the generated data, namely the used velocity field, and the method for obtaining simulated ``radar measurements''.

\paragraph{Velocity field.}
The velocity field in the $xy$-plane was generated from a scalar potential field $\Phi: \mathbb{R}^2 \rightarrow \mathbb{R}$ and the z-component of a vector potential $a: \mathbb{R}^2 \rightarrow \mathbb{R}$. 
Through the Helmholtz decomposition%
\footnote{This is the 2D formulation of the Helmholtz decomposition, where the vector potential has non-zero components only along the $z$-axis as in $\bs a_{\text{3d}} = [0,0,a]^T$. The full decomposition is commonly written as $\bs v_{\text{3d}} = -\nabla \Phi_{\text{3d}} + \nabla \times \bs a_{\text{3d}}$.} %, which in our case reads as $\bs v = -\nabla \Phi + \nabla \times \bs a$, 
we can construct the velocity field $\bs v_{\text{xy}}: \mathbb{R}^2 \rightarrow \mathbb{R}^2$:
\begin{equation}
    \bs v_{\text{xy}}\Bigg(\begin{bmatrix}x \\ y\end{bmatrix}\Bigg) = - \nabla \Phi + \begin{bmatrix}{\delta a}/{\delta y} \\ -{\delta a}/{\delta x}\end{bmatrix}.
\end{equation}
For both experiments the following fields were used:
\begin{align}
    \Phi\Bigg(\begin{bmatrix}x \\ y\end{bmatrix}\Bigg) &= - \frac{1}{2}(x-2) \cdot (y-2),
    \\
    a\Bigg(\begin{bmatrix}x \\ y\end{bmatrix}\Bigg) &= - \frac{1}{5} \exp\Bigg( - \Big( \frac{2}{3}x\Big)^2 - \Big(\frac{2}{3}y\Big)^2 \Bigg).
\end{align}
The derivatives were obtained using the symbolic differentiation library SymPy \citep{sympy}.
To add a nonsteady component, the resulting velocity field is modulated in amplitude as a function of time $t\in [0, 3]$:
\begin{equation}
    \bs v_{\text{xyt}}\Bigg(t, \begin{bmatrix}x \\ y\end{bmatrix}\Bigg)  =  \bs v_{\text{xy}}\Bigg(\begin{bmatrix}x \\ y\end{bmatrix}\Bigg)  \left( \frac{3}{2} \left|\sin\left( \frac{2}{3} \pi t \right)\right| + 0.05\right).
\end{equation}
The $z$ (altitude) component of the velocity only depends on time and is given by:
\begin{equation}
    v_z(t) = 1.6\cdot\sin\left( \frac{4}{3} \pi t \right).
\end{equation}
\paragraph{Simulation.}
For the initial distribution of the fluid, the particle positions were drawn from Gaussian mixtures.
% Starting from time $t=0$ up to $t=3$
For $t\in[0,3]$, these particles were simulated using the above constructed velocity field. 
Overall, the paths of the roughly 240000 parcels were simulated using a basic backward Euler scheme.

\paragraph{Measurements.} The measurements at the sensors were obtained by counting the number of particles within a given radius over multiple timesteps.
The density corresponds to the mass divided by the sensor area, and the velocity is an average over all the particle velocities.
For the training data additional zero-mean isotropic Gaussian noise is added to all measurements.
In the 3D setting, data measurements of density and velocity are obtained by $13^2$ sensors on the $xy$-plane, within region $[-3, 3]^2$ at 11 equidistant timesteps. 
In the 2D setting, the same set of sensors is used.
%the density and velocity are measured spatially less densely, by $15^2$ sensors on the $xy$-plane within the area $[-4, 4]^2$, but temporally more densely at 21 equidistant timesteps

\subsubsection{Architecture and Training}
In both experiments, the networks for density $\rho_{\Theta_1}$ and velocity $v_{\Theta_2}$ prediction (parameterized by ${\Theta_1}$ and ${\Theta_2}$, respectively) are fully-connected layers with sinusoidal activation functions, as proposed by \citet{sitzmann2020implicit}.
The number of layers and units for each setting is shown in Table \ref{table:architecture}.
The sine frequency hyperparameter required in the SIREN architecture was tuned by hand according to the validation loss of the baseline model (i.e. without a PDE loss), leading to a sine-frequency of $12$ for the 2D setting, and $5$ for the 3D setting.
We note that the proposed default value of $30$ in \citet{sitzmann2020implicit} heavily overfits our relatively low-frequency data and we thus recommend an adjustment of this hyperparameter for usage in PINNs.

For training the network, the ADAM optimizer ~\citep{kingma2016adam} with a learning rate of $8\times 10^{-4}$ (2D Setting) or $10^{-4}$ (3D Setting) was used.
The learning rate was multiplied by a factor of $0.99$ each epoch.
All models were trained for 300 (3D setting) or 500 (2D setting) epochs.
The 2D setting was trained using full-batch gradient descent, whereas for the 3D setting we used a mini-batch size of $6931$.
In all experiments we trained and evaluated on 10 different random seeds.

%Experiment &  Input Dimension & \# Hidden Layers & Size Hidden Layers & Output Dimension
\begin{table}[h]
  \caption{Architecture for Particle Simulation Experiments.}
  \label{table:architecture}
  \centering
  \begin{tabular}{llllll}
    \toprule
    %\multicolumn{2}{c}{Part}                   \\
    %\cmidrule(r){1-2}
    Experiment   & Input  & Output Variable     & \# Hidden Layers & \# Hidden Units \\
    \midrule
    2D   & $[0, T]\times \mathbb{R}^2$    & Density\  $\rho_{\Theta_1} \in \mathbb{R}_+$  &  2 & 256  \\
         &    & Velocity $\bs v_{\Theta_2} \in \mathbb{R}^2$ &  1& 64   \\
                 \midrule
    3D    & $[0, T]\times \mathbb{R}^3$   & Density\  $\rho_{\Theta_1}\in \mathbb{R}_+$     &  6 & 256 \\
           &  & Velocity $\bs v_{\Theta_2} \in \mathbb{R}^3$ &    3  & 256 \\
    \bottomrule
  \end{tabular}
\end{table}

\subsection{Experiments: Heat Equation}
\label{appendix:heat}
% Data creation; Details regarding architectures; more plots

%=%=%During the warm-up phase of the pdPINN, collocation points are sampled uniformly, and afterwards $90\%$ of the samples are drawn from the particle density distribution, which is proportional to the modeled temperature.
%=%=%Collocation points are re-sampled every 500 epochs.
%=%=%Differently from previous experiments, the employed architecture is a fully-connected two-layer neural network with 32 hidden units and tanh activations.
%=%=%As in previous settings, we show that with few samples the regularization enforced by the PDE loss is not strong enough, leading to comparable results in both approaches.
%=%=%However, as the number of sample increases, the PDE loss enforced with MCMC samples quickly and steadily outperforms uniform sampling.
%=%=%We also verified that in the limit of high samples the two sampling strategies converge, as expected.
The dataset for the heat equation experiment was generated by numerically solving the heat equation through the finite difference method, precisely the Forward Time, Centered Space (FTCS) approximation~\citep{Recktenwald2004FiniteDifferenceAT}.
%This requires including boundary conditions, which in this case were set 
We used Dirichlet boundary conditions in form of zero temperature around a squared shape far away from the relevant domain.
These boundary conditions are not provided to the PINNs for a slightly more difficult setting.
Overall, the dataset is composed of 1000 training points, 1971120 test points and 492780 validation points. 
We made sure training points contained enough information about the initial condition, i.e. we selected a sufficient amount of points around the initial source of non-zero temperature.
In contrast, validation and test points are taken uniformly in time and space.
During the warm-up phase of the pdPINN training, collocation points were sampled uniformly, and afterwards $90\%$ of the samples were drawn from the particle density distribution, which is proportional to the modeled temperature.
Collocation points were re-sampled every 500 epochs.
Differently from previous experiments, the employed architecture is a fully-connected two-layer neural network with 32 hidden units and tanh activations.
The implementation is in PyTorch \citep{pytorch}, using the ADAM optimizer ~\citep{kingma2016adam} combined with an exponential learning rate scheduler which multiplies the learning rate by a factor of $0.9999$ at each epoch, starting with a rate of $10^{-4}$ and decreasing it until reaching a minimum value of $10^{-5}$.
Training was terminated through early-stopping, as soon as the validation R$^2$ didn't improve for more than 3000 epochs.

\begin{figure}[ht!]
      \centering
      \includegraphics[width=1\linewidth]{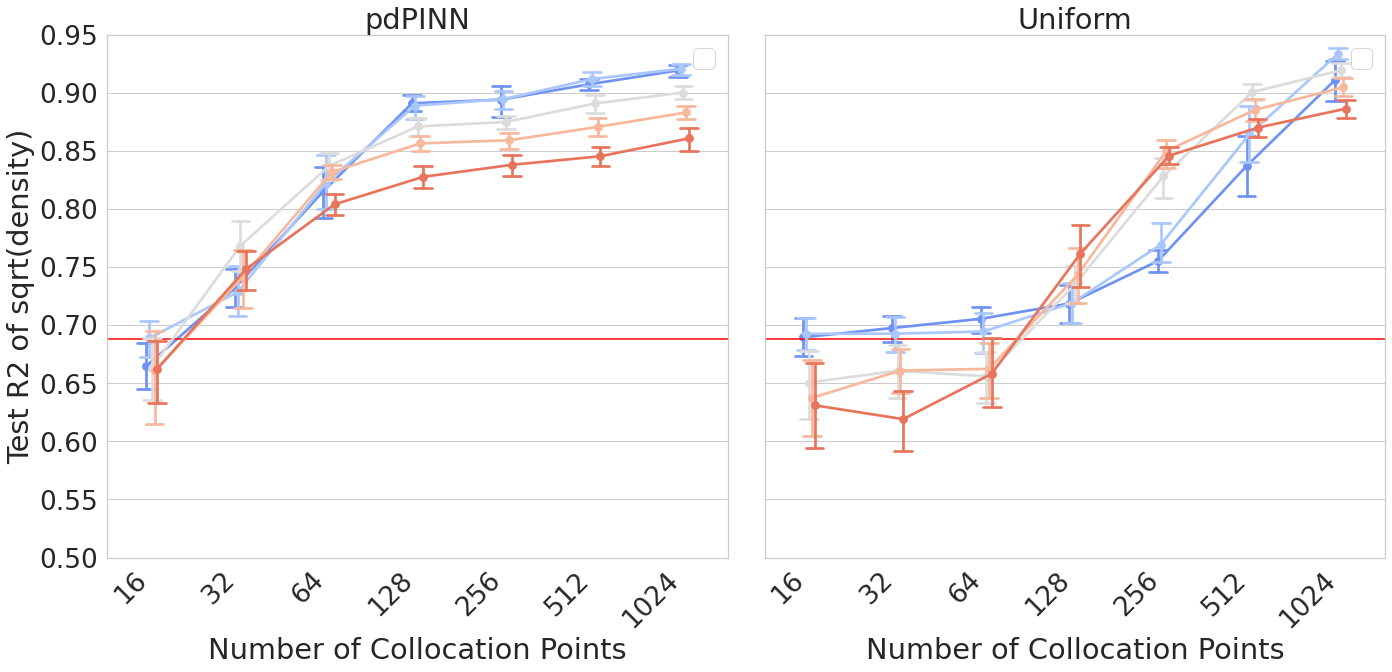}
      \caption{Test R$^2$ of predicted $T$ in the heat equation experiment as a function of different number of collocation points. 
      Results are averaged over 20 different seeds and the resulting error bars correspond to $95\%$ confidence interval for the mean estimation, based on 1000 bootstrap samples.
      Different colors indicate different PDE weights $w_2$.
      }
      \label{fig:expheat_r2_results_weights}
      %PDE weights in the range $w_2\in\{0.005,0.01,0.05,0.1,0.2\}$ are illustrated with different colors shades \hl{from blue to red} }
\end{figure}

\paragraph{Additional results.}
Figure \ref{fig:expheat_r2_results_weights} illustrates the test R$^2$ of the predicted $T$ averaged over 20 different seeds. 
Error bars correspond to $95\%$ confidence interval for the mean estimation, based on 1000 bootstrap samples, while colors indicate the different PDE weights $w_2$ explored.
As in previous settings, we show that with few samples (16) the regularization enforced by the PDE loss is not strong enough, leading to comparable results in both approaches (as expected).
%When the PDE loss is enforced with few samples (16) the regularization is not strong enough, which is to be expected.
Hence PINNs and pdPINNs show similar results in this regime.
However, as the number of samples increases (32-64-128-256), the PDE loss enforced by the proposed pdPINNs quickly and steadily outperforms uniform sampling.
%Lastly, for a sufficiently large number of samples (512-1024) we see no significant difference within the two sampling procedures,
Lastly, we also verified that in the limit of high samples (512-1024) the two sampling strategies converge, as in such a low-dimensional domain the uniform samples fully and densely covers the considered area.
This, again, is in line with the observed results of the other experiments.
%For the sake of completeness we illustrate results for each PDE weight loss explored in Figure \ref{fig:expheat_r2_results_weights}.
%The explored weights are over the range $w_2\in\{0.005,0.01,0.05,0.1,0.2\}$.

\subsection{Experiments: Fokker-Planck Equations in TensorFlow}
\label{appendix:fp}
%For an additional problem, i.e. a setting without any observed data available, we consider the Fokker-Planck equation.
Within the Fokker-Planck experiment we showcase the different training behaviors of uniform sampling, RAR, and multiple MCMC samplers.
Due to the low dimensionality of the problem, we additionally consider a Inverse-Transform (IT) sampler \citep{steele1987non} for efficiently sampling from the density.
The IT sampler relies on the empirical cdf estimated via uniform samples drawn over the whole domain.
This method does not require building up a Markov Chain, and is thus very fast, but only works well in low dimensions.

More specifically, we compare the following methods for selecting collocation points, with a highly efficient implementation of the MCMC methods provided by TensorFlow probability:
\begin{enumerate}[I.)]
    \item Uniform sampling
    \item Residual Adaptive Refinement \citep{lu2021deepxde}
    \item pdPINN with Inverse-Transform (IT) sampling  \citep{steele1987non}
    \item pdPINN with Metropolis-Hastings (MH) MC with parallel tempering \citep{earl2005parallel}
    \item pdPINN with Hamiltonian MC (HMC) with parallel tempering \citep{earl2005parallel} and dual averaging step-size adaptation \citep[section 3.2]{hoffman2014no}
\end{enumerate}

%This experiment is done with an additional TensorFlow implementation, leveraging the efficient MCMC implementations provided by TensorFlow probability \citep{abadi2016tensorflow, lao2020tfp}. 
%Furthermore, we make use of the DeepXDE library \citep{lu2021deepxde}.
%We note, that due to the software design of DeepXDE the implemented RAR method in this experiment is much slower and not representative - an efficient implementation should be as fast as the shown Inverse Transform Sampler and the Uniform sampler.

\subsubsection{Setting and Analytical Solution}
% The Fokker-Planck equations describe the evolution of the probability density of the movement of Brownian particles under a drift.
% \begin{figure}[h!]
% \centering\includegraphics[width=.6\textwidth]{images/exp_fp/fp_1d_uniform_20ks_50kit.png}
% \caption{Solution of the 1D Fokker-Planck equation obtained via a PINN with uniform samples.}
% \label{figure:fp_1d_img}
% \end{figure}

% More specifically, assume we are given particles at time $t_0$, which are distributed according to $p(x,t_0)$.
% Let the movements of these particles be described by the following stochastic differential equation, where $W_t$ denotes the standard Wiener process:
% $$ dX_t = \mu(X_t, t) \,dt + \sigma(X_t, t) \,dW_t $$

% with drift $\mu(X_t, t)$ and diffusion coefficient $D(X_t, t) = \sigma^2(X_t, t)/2$, the Fokker–Planck equation for the probability density $p(x, t)$ of the random variable $X_t$ is
% $$
% \frac{\partial}{\partial t} p(x, t) = -\frac{\partial}{\partial x}\left[\mu(x, t) p(x, t)\right] + \frac{\partial^2}{\partial x^2}\left[D(x, t) p(x, t)\right] .
% $$
We consider the following setting over the time interval  $[t_0, t_n] = [-1,1]$ with  
drift function ${\mu}$, noise ${\sigma}$ and initial particle positions $p(x|t=t_0)$ given by
\begin{align}
\mu({X}_t,t) &= {\mu}(t) = \sin\left(10 t\right)\\
{\sigma}({X}_t,t) &= \sigma =0.06\\
p(x|t=t_0)&=\mathcal{N}(0, 0.02^2 \cdot I_d )
\end{align}

The PDE has an analytical solution (cf. \cite{sarkka2019applied}) which is given by 
\begin{align}
    p(x|t) &= \mathcal{N}(\mu_s(t), \sigma^2_s(t)) 
    \\
    p(t) &= \mathcal{U}(t_0, t_n)
    \\
    \mu_s(t) &= -\frac{\cos(10t)}{10} + \frac{\cos(10)}{10}\\
    \sigma^2_s(t) &= 0.0036t + 0.004 .
\end{align}
For evaluating the deviation of our prediction to the solution, we evaluate the KL divergence between the analytical solution and the network approximation $KL(p(x,t)|\hat{p}_\Theta(x,t))$ by sampling 10000 points from the true $p(x,t)$.
% We furthermore consider the extension to a 2D setting, with the same drift function and noise in each dimension:
% \begin{equation}
%     \frac{\partial p(\mathbf{x},t)}{\partial t} = -\sum_{i=1}^N \frac{\partial}{\partial x_i} \left[ \mu_i(\mathbf{x},t) p(\mathbf{x},t) \right] + \sum_{i=1}^{N} \sum_{j=1}^{N} \frac{\partial^2}{\partial x_i \, \partial x_j} \left[ D_{ij}(\mathbf{x},t) p(\mathbf{x},t) \right] ,
% \end{equation}
% with
% \begin{equation}
%     \mu_i(\boldsymbol{x},t) = \mu(t)
% \end{equation}
% \begin{equation}
%     D_{ij}(\mathbf{x},t) = \frac{1}{2}\sum_{k=1}^M \sigma_{ik}(\mathbf{x},t) \sigma_{jk}(\mathbf{x},t) \qquad \text{ where } \sigma_{jk}(\boldsymbol{x}, t) = 
%     \begin{cases}
%       \sigma, & \text{if}\ i = j \\
%       0 & \text{else}
%     \end{cases}
% \end{equation}

\subsubsection{Setup}
%For both the 1D and 2D setting 
We use a SIREN network and additionally sample (5000) collocation points at the initial time-step, which is the default behavior of DeepXDE.
An overview of the architecture and training details is given in Table \ref{table:architecture_fp}.
Experiments were performed with a NVIDIA GeForce RTX 2080 Ti and an Intel(R) Xeon(R) CPU E5-1660 v3 @ 3.00GHz processor.

%Experiment &  Input Dimension & \# Hidden Layers & Size Hidden Layers & Output Dimension
\begin{table}[h!]
  \caption{Architecture for Fokker-Planck experiments.}
  \label{table:architecture_fp}
  \centering
  \begin{tabular}{lllllll}
    \toprule
    %\multicolumn{2}{c}{Part}                   \\
    %\cmidrule(r){1-2}
    Experiment   & Input  & Output Variable     & \# Layers & \# Units & col. points & epochs \\
    \midrule
    1D   & $[0, T]\times \mathbb{R}$    &   $p_{\Theta}(x,t) \in \mathbb{R}_+$  &  5 & 64 & 5.000 & 30.000 \\
                 %\midrule
    %2D    & $[0, T]\times \mathbb{R}^2$   &   $\hat{p}_{\Theta}(\boldsymbol{x},t)\in \mathbb{R}_+$     &  9 & 128 & 30.000 & 30.000 \\
    \bottomrule
  \end{tabular}
\end{table}

\subsubsection{Wall time}
%The results in Figure \ref{figure:fp_1d} show that all methods lead to a low KL divergence with sufficient training steps.
%With respect to the number of steps (or epochs), all pdPINN sampling methods seem to be preferable or at least comparable to RAR and uniform sampling in their speed of convergence.
%However, taking a closer look at the wall-time, it can be seen that Metropolis-Hastings and Hamiltonian Monte Carlo require significantly more time, whereas the inverse transform sampling is comparable to uniform sampling.
%We again highlight that a more efficient RAR implementation would show similar runtimes to uniform sampling.

\begin{figure}[h]
\centering\includegraphics[width=.5\textwidth]{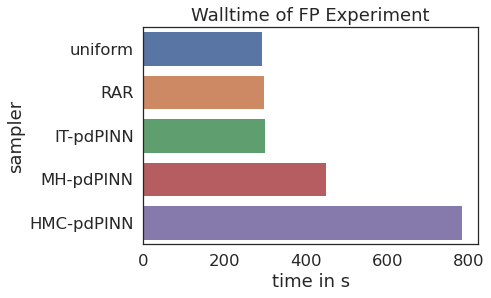}
\caption{Total run-times for the Fokker-Planck experiment. Seeds were selected randomly.}
\label{figure:fp_walltime}
\end{figure}

%This shows that the specific sampling method has to be selected based on the problem at hand.
The wall times for the different methods are provided in Figure \ref{figure:fp_walltime}.
Although Metropolis-Hastings and Hamiltonian Monte Carlo require more time per step compared to uniform sampling, the used inverse transform sampling achieves a similar speed.
%This showseven in low-dimensional settings where Uniform sampling is sufficient, a fast sampling method for pdPINNs might be preferable.

% \begin{figure}[h!]
% \centering\includegraphics[width=.9\textwidth]{images/exp_fp/comparison_1d.png}
% \caption{KL divergence and total run-times for the 1D Fokker-Planck experiment. The shown KL divergences are a rolling average with a window of 4000 epochs. Seeds were selected randomly.}
% \label{figure:fp_1d}
% \end{figure}

% \subsubsection{Results in 2D}
% A slightly more difficult problem is given by the 2D Fokker-Planck equation.
% Although in the previous setting simple samplers were sufficient, sampling in 2+1 dimensions already requires more sophisticated methods, as can be seen in Figure \ref{figure:fp_2d}.
% All methods relying on uniform proposals either completely failed, or showed great instabilities during training.
% Furthermore, due to the larger amount of collocation points, the computational cost of sampling becomes negligible compared to computing the higher order derivatives for the PDE.
% Qualitative results of the prediction after 30000 epochs are shown in Figure \ref{figure:fp_2d_img}

%\subsection{Performance for different number of Collocation Points}
%\label{appendix:collocation}

%\begin{figure}[ht!]
%      \centering
%         \includegraphics[width=.99\linewidth]{images/overview_nsamples_pdeweight/exp_heateq_nsamples_all_weights_boxplot.png}
%      \caption{Heat Equation for  different number of collocation points. The colors indicate different PDE weights.\hl{From left to right: ,,,,}}
%      \label{app:fig:exp2d_n_samples_all_weights_boxplot}
%\end{figure}

% \begin{figure}
%     \centering
%     \includegraphics[width=0.99\linewidth]{images/exp3d/exp3d_n_samples_all_weights.png}
%     \caption{All combinations of weights and number of collocation points for the 3D experiment. The colors indicate different PDE-weights.}
%     \label{fig:my_label}
% \end{figure}

%\begin{figure}
%  \centering
%  
%      \centering
%        \includegraphics[width=.3\linewidth]{images/exp_heatequation/exp_heateq_nsamples.png}
%      \label{fig:heat_eq_nsamples}
%  \caption{Heat Equation}
%\end{figure}

\end{document}